\documentclass{article}
\usepackage{amsmath}
\usepackage{pdfpages}
\usepackage{graphicx}
\usepackage{subcaption}
\usepackage{booktabs}
\usepackage{pifont}
\usepackage[table]{xcolor}
\usepackage{tcolorbox}
\usepackage{fvextra}
\usepackage{enumitem}
\usepackage{hyperref}
\usepackage{microtype}
\usepackage{bbm}
\usepackage{longtable}
\usepackage{placeins}
\usepackage{tabularx}
\tcbuselibrary{breakable}
\hypersetup{
    colorlinks=true,
    linkcolor=blue,
    urlcolor=blue,
    pdfborderstyle={/S/U/W 0}
}

\title{AMIGO: Agentic Multi-Image Grounding Oracle Benchmark}
\author{Min Wang,  Ata Mahjoubfar\\
Target Corporation\\
{\tt\small \{Min.Wang, Ata.Mahjoubfar\}@target.com}}

\date{}
\begin{document}
\maketitle

\begingroup
\renewcommand{\thefootnote}{}
\footnotetext{The dataset is provided as an ancillary file, \texttt{anc/data.json}.}
\endgroup
\begin{abstract}

 Agentic vision-language models increasingly act through extended interactions, but most evaluations still focus on single-image, single-turn correctness. We introduce \textbf{AMIGO} (\textbf{A}gentic \textbf{M}ulti-\textbf{I}mage \textbf{G}rounding \textbf{O}racle Benchmark), a long-horizon benchmark for \emph{hidden-target} identification over galleries of visually similar images. In AMIGO, the oracle privately selects a target image, and the model must recover it by asking a sequence of attribute-focused Yes/No questions under a strict protocol that returns Yes/No/Unsure feedback and penalizes invalid actions with \emph{Skip}. This setting stresses (i) question selection under uncertainty, (ii) consistent constraint tracking across turns, and (iii) fine-grained discrimination as evidence accumulates. We instantiate AMIGO with the \textit{Guess My Preferred Dress} task and evaluate open-source VLMs with metrics covering identification success, evidence verification, efficiency, protocol compliance, robustness to controlled feedback perturbations, and trajectory-level diagnostics. The benchmarking results show that final-answer accuracy alone overstates evidence-grounded performance: models can guess correctly without verification-passing evidence, waste turns through invalid questions, or fail to preserve the upload protocol. Strong AMIGO performance depends on the combination of visual discrimination, informative question selection, constraint tracking, efficient stopping, sustained protocol following, and recovery from controlled feedback noise; model scale alone does not guarantee reliable long-horizon interactive grounding.
\end{abstract}
\section{Introduction}
Most multimodal benchmarks present the relevant visual context upfront and score a final response. This leaves an important gap: many practical vision-language model (VLM) interactions require the model to reduce uncertainty before answering. In these settings, the model must decide what visual evidence is missing, ask targeted questions, incorporate user feedback, and delay its final action until the accumulated constraints are sufficient. These abilities are difficult to measure in static, short-horizon evaluations~\cite{vqav2,infovqa,mmmu,mmbench,mme}, where strategy, memory, and belief revision are mostly invisible.

We study a complementary setting: interactive hidden-target identification over image collections. A user (or oracle) privately selects a target image from a gallery of visually similar candidates, and the model must discover the target by asking constrained questions about observable attributes. Because the target is hidden from the model but fixed for the oracle, success requires the model to construct an evidence trail rather than merely recognize a salient object from a prompt. This formulation surfaces agentic challenges that are often invisible in single-turn evaluation: choosing questions that efficiently reduce ambiguity, preserving and applying constraints without drift, and performing cross-image comparisons that depend on the evolving candidate set rather than any single image in isolation.

To operationalize this setting, we present AMIGO, an interactive benchmark centered on a simple but diagnostic loop. The model asks exactly one Yes/No question per turn, receives Yes/No/Unsure feedback, and iteratively narrows the feasible candidate set. The interaction is bounded by a fixed question budget, making efficiency and invalid actions directly comparable across models. If the model violates protocol rules, the oracle returns \emph{Skip}, revealing no information. Because each invalid action consumes a turn while providing no attribute evidence, \emph{Skip} exposes failures that final-answer metrics miss: models may appear concise or confident while losing opportunities to gather useful information. AMIGO's verification module further separates three outcomes that are often conflated: incorrect identification, correct final guesses without sufficient supporting evidence, and Verified Correct trajectories that narrow the feasible set to the hidden target. This distinction is central for evaluating agentic grounding, where a plausible final answer is less informative than the interaction path that justifies it.

We implement AMIGO through the \textit{Guess My Preferred Dress} task, where each gallery contains 6 to 40+ highly similar dress images. Dresses offer rich fine-grained variation (e.g., neckline construction, closures, seam placement, drape, and embellishments), making the task sensitive to careful observation and long-horizon constraint tracking. The domain is also well suited to controlled protocol design: highly salient shortcut attributes such as color, pattern, sleeve length, and garment length can be prohibited, forcing models to rely on subtler visual evidence.

\noindent\textbf{Contributions.}
(1) We introduce AMIGO, a benchmark that combines full-image multi-candidate comparison, constrained attribute questioning, explicit invalid-action feedback, and model-relative trajectory verification for evidence-grounded hidden-target identification over visually similar image galleries.
(2) We define a constrained Yes/No question protocol with Yes/No/Unsure oracle feedback and explicit invalid-action feedback (\emph{Skip}), making invalid actions and wasted interaction budget directly measurable.
(3) We propose verification-based metrics that separate final-answer correctness from evidence-grounded success and expose correct guesses that lack verification-passing support.
(4) We benchmark open-source VLMs and observe that strong performance depends on more than model scale: models must combine fine-grained visual discrimination, informative question selection, constraint tracking, efficient stopping, and sustained protocol following.
(5) We present structured trajectory logs and diagnostics, including a controlled noisy-feedback robustness test, that support failure analysis and can serve as a foundation for future reinforcement learning or preference-optimization studies of long-horizon multimodal policies.

\noindent\textbf{Implications.}
AMIGO makes long-horizon grounding behavior measurable rather than reducing evaluation to a final answer. Its logs show how models choose questions, update candidate sets, handle ambiguous or perturbed feedback, and decide when to stop. This framing also makes calibration visible: a reliable model should keep asking valid questions when evidence is insufficient instead of committing early.

\section{Related Work}

\subsection{Multi-image and multi-turn multimodal evaluation}
A growing set of benchmarks and datasets extends evaluation beyond single-image, single-turn QA~\cite{mmmu, infovqa} to multi-image and/or multi-turn settings. Multi-image benchmarks (e.g., MMIU~\cite{mmiu}, MuirBench~\cite{muirbench}, mPLUG-Owl3~\cite{mplugowl3}) probe cross-image grounding and long visual-context handling. Multi-image instruction-tuning datasets (e.g., Mantis~\cite{mantis}) and multi-image multi-turn resources (e.g., MMDU~\cite{mmdu}, MMCR~\cite{mmcr}) further study supervision over interleaved images and dialogue. Separately, multi-turn multimodal conversation benchmarks (e.g., MultiVerse~\cite{multiverse}, ConvBench~\cite{convbench}, MMMT-IF~\cite{mmmtif}, MMCoQA~\cite{mmcoqa}) evaluate contextual coherence and instruction adherence over longer dialogues.

These efforts substantially expand visual context length and dialogue length, but the model is still typically evaluated as a responder to provided inputs. AMIGO instead evaluates interactive hidden-target grounding: the model must choose discriminative visual attributes, update constraints from oracle feedback, and stop only when the accumulated evidence supports a unique target.

\subsection{Multimodal reasoning benchmarks}
General multimodal reasoning benchmarks assess whether models can integrate vision and language for multi-step inference beyond shallow cue matching. For example, EMMA~\cite{emma} targets ``organic'' multimodal reasoning across domains, and MLLM-CompBench~\cite{mllmcompbench} emphasizes comparative reasoning under multimodal inputs. MMMU-Pro~\cite{mmmupro} further strengthens this line by making evaluation more resistant to text-only shortcuts and by requiring tighter visual grounding in challenging multimodal questions.

These benchmarks are valuable for measuring perception-and-inference capability, but the reasoning problem is usually fully specified by the prompt and visual input. AMIGO makes part of the problem state hidden, so reasoning quality is expressed through the model's question policy and evidence-accumulation process, not only through its final answer.

\subsection{Interactive Visual Grounding and Target Identification}

AMIGO is closely related to interactive settings in which a system must resolve hidden or underspecified task state through dialogue before making a decision. Recent work studies this problem in the context of language-model agents: LMRLGym~\cite{lmrl} benchmarks reinforcement learning for multi-turn language-model agents across interactive dialogue and text-game environments, including a Twenty Questions task that requires strategic questions to identify a hidden entity. This formulation highlights long-horizon decision making, credit assignment, and action selection that improves eventual task success rather than immediate response quality. InfoQuest~\cite{infoquest} evaluates text-only models that ask clarifying questions when critical context is hidden. ClariMM~\cite{clarimm} extends this idea to multimodal clarification for underspecified user queries. More broadly, multimodal agent frameworks such as M3Searcher~\cite{m3searcher} and expert-guided benchmarks such as MIRAGE~\cite{mirage} follow a ``seek-information-then-decide'' paradigm, often with retrieval or tool-mediated evidence acquisition.

A closely related line of earlier work studies visually grounded guessing games with task-specific dialogue models. GuessWhat~\cite{guesswhat} introduced a two-player task in which a questioner identifies a hidden object within an image by asking a sequence of questions answered by an oracle. Das et al.~\cite{das2017learning} formulated a cooperative visual-dialog image-guessing game in which Q-BOT must recover an unseen image from a candidate lineup through multi-round interaction with A-BOT, training both dialogue models using deep reinforcement learning. Chattopadhyay et al.~\cite{chattopadhyay2017evaluating} adapted this setting into the human--AI GuessWhich game, evaluating whether a person can identify the secret image by conversing with a visual-dialog model. Later work developed attentive-memory architectures~\cite{guesswhich} and multimodal dialogue-state tracking~\cite{pang2023multi} for this general image-guessing paradigm.

Interactive visual-grounding benchmarks provide another closely related direction. InViG~\cite{invig} requires an agent to resolve an ambiguous object reference by asking clarifying questions and incorporating user feedback. PhotoBook~\cite{photobook} studies multi-image conversational grounding, repeated reference, and the construction of common ground between interlocutors. Together, these works establish interactive visual identification as a test of grounded language understanding, dialogue-history modeling, and referent resolution.

Whereas prior guessing-game and interactive-grounding benchmarks focus primarily on task success, AMIGO makes the interaction policy itself an explicit object of evaluation. GuessWhat-style tasks identify a hidden object within a single image, while GuessWhich and related visual-dialog games are typically framed as recovering or identifying an image through broader free-form dialogue. InViG focuses on resolving ambiguous object references through clarification, and PhotoBook studies common-ground construction across repeated visual references. AMIGO instead fixes a closed gallery of highly similar full-image candidates and restricts the model to one binary, attribute-focused question per turn. This design exposes failures that are difficult to measure in less constrained settings: invalid attribute choices, redundant question enumeration, premature guessing, upload-phase protocol violations, candidate-set stagnation, and correct final guesses that lack verification-passing evidence. AMIGO scores the path to the answer directly: whether each question is valid and informative, whether constraints shrink the candidate set, and whether the final guess is supported by the accumulated evidence. The resulting logs provide a well-defined action, observation, and candidate-state record for trajectory analysis.

\section{Benchmark}
\begin{figure}[t]
    \centering
    \includegraphics[width=\linewidth]{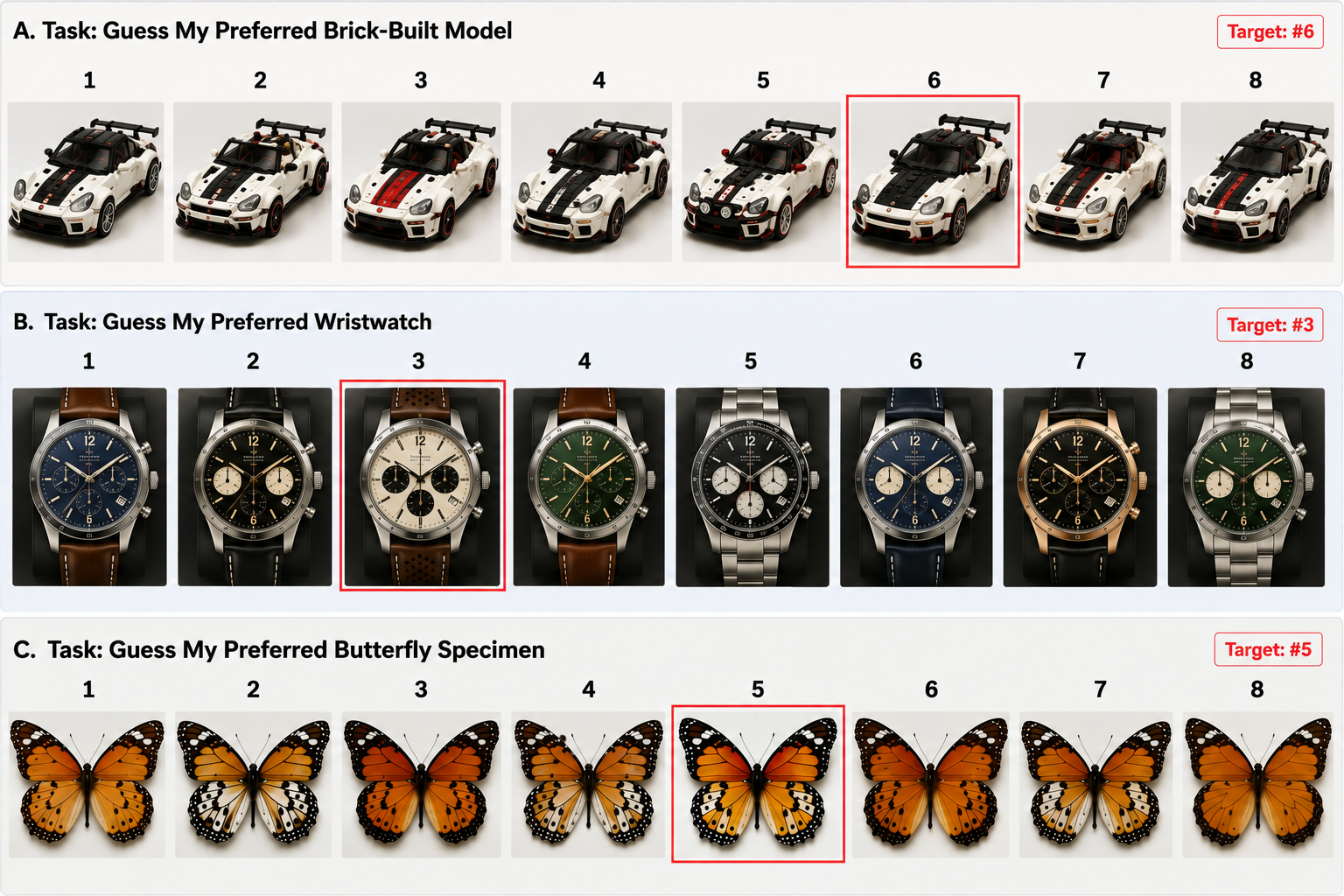}
    \caption{Candidate galleries for AMIGO task extensions beyond dresses; red outlines mark selected targets.}
    \label{fig:alternative_tasks}
\end{figure}

We introduce AMIGO, a benchmark designed to evaluate agentic vision-language models on hidden-target identification in multi-image, multi-turn settings. AMIGO operationalizes this paradigm through its primary task, \textit{Guess My Preferred Dress}, in which a model must identify a privately selected target dress from a gallery of highly similar candidates by asking discriminative, attribute-focused questions under a strict interaction protocol. Beyond this primary instantiation, the same protocol could extend to other closed-gallery domains with rich fine-grained variation. Representative examples include \textit{Guess My Preferred Brick-Built Model}, \textit{Guess My Preferred Wristwatch}, and \textit{Guess My Preferred Butterfly Specimen}, which respectively emphasize structural, manufactured, and natural visual evidence; Figure~\ref{fig:alternative_tasks} shows example candidate galleries for these extensions.

\subsection{Guess My Preferred Dress}
\paragraph{Task.}
A user uploads one or more batches of dress images that together form a single gallery, with one image privately selected by the user as the hidden target. After the signal \texttt{End of uploading}, the model asks at most 20 questions to narrow the candidate set and finally outputs the 1-indexed position of the target in the gallery.

\paragraph{Protocol.}
The interaction is governed by a constrained question-answer protocol. These rules define the protocol-compliance rubric used to classify invalid actions and interpret \emph{Skip} feedback:
\begin{enumerate}[leftmargin=*, label=\textbf{P\arabic*.}]
  \item \textbf{Question format.}
  Each turn must contain exactly one Yes/No question about one observable attribute of the target dress.

  \item \textbf{Oracle response.}
  The question must be answerable with \textbf{Yes}, \textbf{No}, or \textbf{Unsure}.

  \item \textbf{Invalid-action feedback.}
  If the model violates any protocol rule, the oracle returns \emph{Skip}; a \emph{Skip} response reveals no attribute information.

  \item \textbf{Prohibited attributes.}
  The model must not ask about sleeve length, garment length, piece count, color, prints/patterns, age group, size, shoes, necklace, hat, bag, background, or the human model.

  \item \textbf{Redundancy constraint.}
  The model must not enumerate values of the same attribute type across turns, such as repeatedly cycling through neckline types (e.g., V-neck, square neck, crewneck).

  \item \textbf{Index-reference constraint.}
  The model must not reference specific image indices or inspect images one by one (e.g., ``Is it image \#1?'').

  \item \textbf{Stopping rule.}
  The model should not guess until its accumulated constraints narrow the feasible candidate set to exactly one image.

  \item \textbf{Verification under uncertainty.}
  If constraints appear inconsistent or uncertain, the model should continue asking verification questions rather than stopping early.

  \item \textbf{Final-answer format.}
  When ready, the model must output: \texttt{My guess of your favorite dress: \#<number>.}

  \item \textbf{Upload phase.}
  The model must not ask questions until the user sends \texttt{End of uploading}. Echoing the upload-completion phrase is not counted as a P10 violation unless the output includes a question.

\item \textbf{Question budget.}
After the \texttt{End of uploading} signal, the model may ask at most 20 questions before making a final guess. If the model exhausts this budget without producing a final guess, the trajectory terminates and is counted as incorrect. Questions generated during the upload phase are recorded as premature outputs but do not count toward this budget.

\end{enumerate}

\paragraph{Rationale.}
The protocol deliberately excludes highly salient cues (e.g., color, patterns, piece count, sleeve length, and garment length) to encourage reliance on subtler construction details (e.g., neckline structure, closures, waist shaping, seam placement, drape, and embellishments). This design emphasizes long-horizon constraint tracking and rewards question sequences that are informative, non-redundant, and robust to uncertainty.

\paragraph{Example interactions.}
The example trajectories in Figure~\ref{fig:conversation_examples} illustrate how visually similar candidates can lead to different outcomes under the same protocol. The failed trace in Figure~\ref{fig:conversation_fail} contains useful construction-focused questions but still leaves confusable candidates unresolved, whereas the successful trace in Figure~\ref{fig:conversation_correct} uses a more discriminative sequence to isolate the hidden target.

\begin{figure}[htbp]
  \centering
  \begin{subfigure}[t]{\textwidth}
    \centering
    \includegraphics[width=\textwidth]{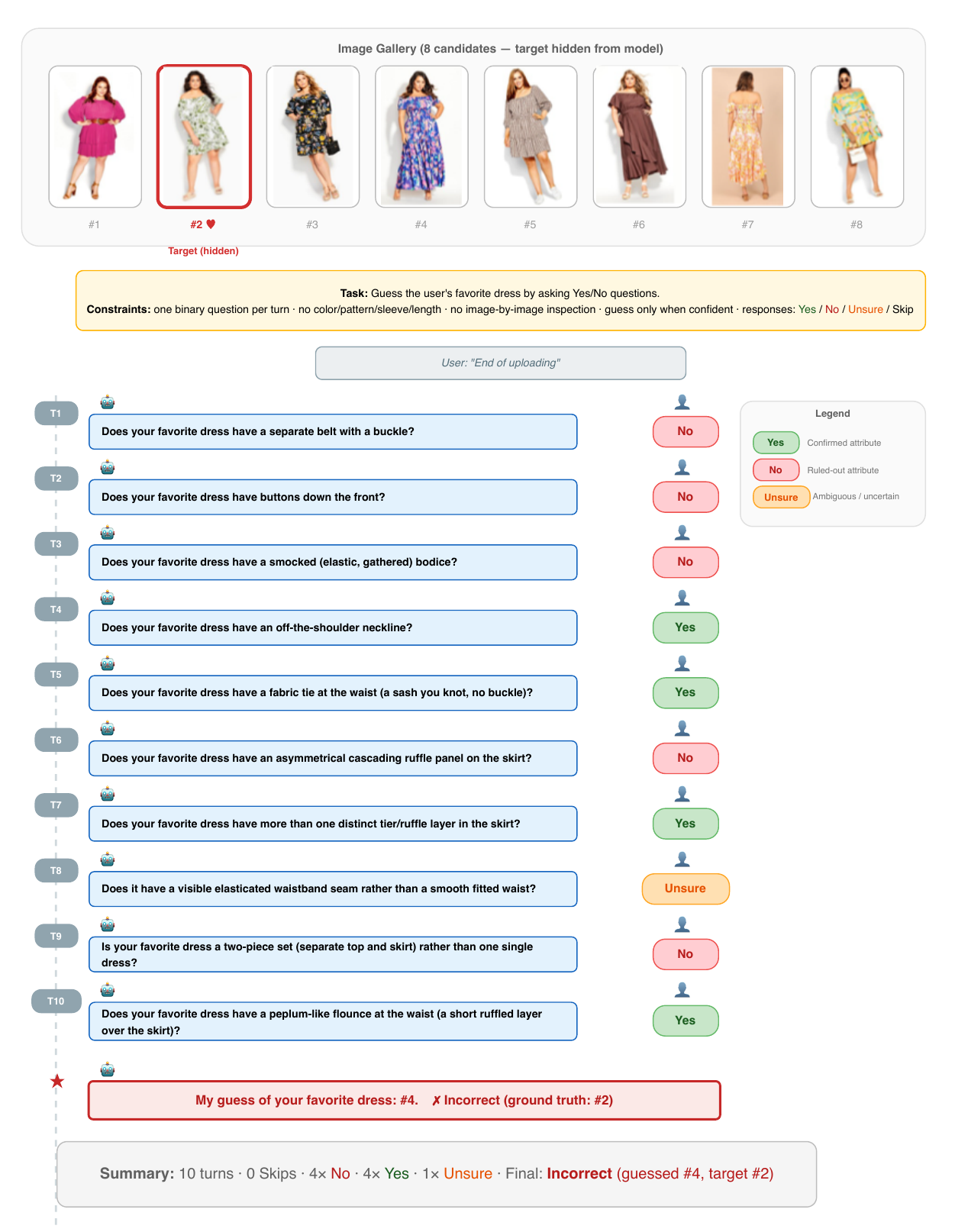}
    \caption{Failed trajectory: construction-focused questions do not resolve the visually similar candidates.}
    \label{fig:conversation_fail}
  \end{subfigure}
  \caption{Two AMIGO interactions for \textit{Guess My Preferred Dress}, contrasting (a) an incorrect final guess with (b) a successful target identification.}
  \label{fig:conversation_examples}
\end{figure}

\begin{figure}[htbp]
  \ContinuedFloat
  \centering
  \begin{subfigure}[t]{\textwidth}
    \centering
    \includegraphics[width=\textwidth]{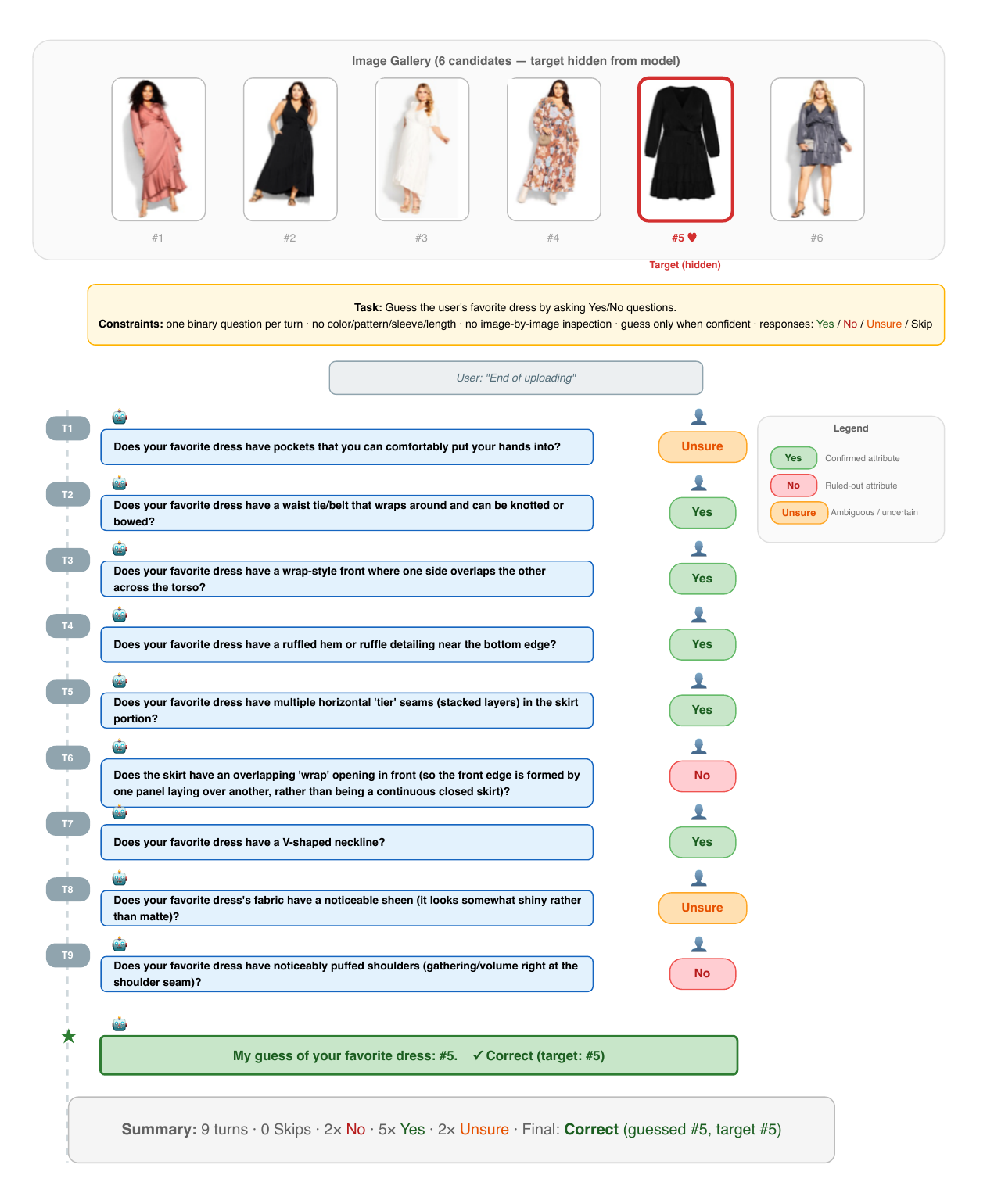}
    \caption{Successful trajectory: a discriminative nine-turn exchange identifies the hidden target.}
    \label{fig:conversation_correct}
  \end{subfigure}
  \caption[]{Two AMIGO interactions for \textit{Guess My Preferred Dress} (\emph{continued}).}
\end{figure}

\subsection{Dataset collection}
\paragraph{Source and curation.}
We collect dress images from an online retailer's fashion catalog and deduplicate them to ensure each product is unique. Attribute metadata extracted from the retailer's historical product database is often missing, defined inconsistently across brands, and sometimes contains erroneous or irrelevant values. We therefore prioritize visually verifiable attributes and build a semi-automatic pipeline to produce standardized labels at scale.

\begin{figure}[htbp]
  \centering
  \includegraphics[width=\textwidth]{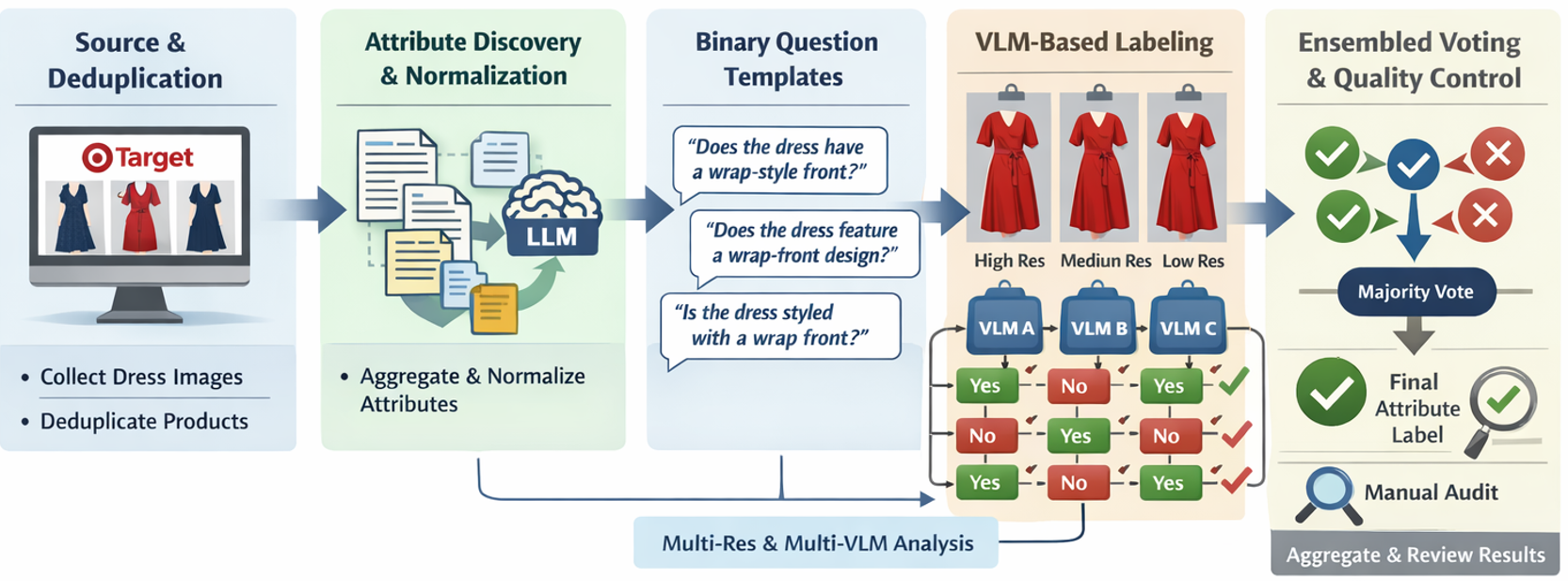}
  \caption{Semi-automatic attribute labeling pipeline that normalizes catalog attributes, generates binary question templates, and ensembles VLM responses into standardized labels.}
  \label{fig:attributes_labeling_pipeline}
\end{figure}

\paragraph{Semi-automatic attribute labeling.}
Our pipeline has three stages:
(1) \textbf{Attribute discovery and normalization:} we aggregate attribute values from catalog annotations and use multiple LLMs to normalize synonyms and naming variants, compile a comprehensive attribute value list, and expand it with as many visually grounded alternatives as possible;
(2) \textbf{Binary question template construction:} for each normalized attribute value, we generate multiple binary (Yes/No) question templates (e.g., ``Does the dress have a wrap-style front?'', ``Does the dress have a wrap-front design?'', ``Does the dress feature a wrap-style front?'');
(3) \textbf{VLM-based labeling:} we apply multiple VLMs to answer these templates for each image, yielding per-image attribute labels.
Figure~\ref{fig:attributes_labeling_pipeline} summarizes this semi-automatic labeling pipeline.

\paragraph{Ensemble VLM labeling and quality control.}
Given budget constraints, we adopt open-source VLMs for automated labeling. To improve labeling reliability, we ensemble multiple VLMs (\textbf{Qwen3-VL-235B-A22B-Instruct-FP8}~\cite{qwen3}, \textbf{Intern-S1}~\cite{intern-s1}, \textbf{GLM-4.5V}~\cite{glm45}) and combine their predictions through weighted majority voting. In our experiments, \textbf{Qwen3-VL-235B-A22B-Instruct-FP8} achieved the best performance among the models evaluated prior to the release of \textbf{Qwen3.5-397B-A17B-FP8}, and is therefore assigned a larger voting weight. We further enhance robustness by (i) using paraphrased question templates to reduce sensitivity to prompt phrasing and (ii) applying multi-resolution image augmentation to mitigate failures in recognizing small visual details. Finally, we manually audit a subset of the labels to assess quality, identify systematic errors, and refine the templates and normalization rules accordingly.

\paragraph{Attribute-based similarity.}
Let $Attr(X)$ denote the set of attribute values assigned to an image $X$. We define an asymmetric similarity score:
\begin{equation}
\text{Sim}(A, B) = \frac{|Attr(A) \cap Attr(B)|}{|Attr(A)|},
\end{equation}
and analogously $\text{Sim}(B, A)$. We rank candidates $B$ by $\mathrm{Sim}(A,B)$ to retrieve candidate images that are most visually similar to the reference image $A$ in terms of attribute overlap (i.e., candidates that best cover $A$'s attributes).

\begin{figure}[htbp]
  \centering
  \includegraphics[width=\textwidth]{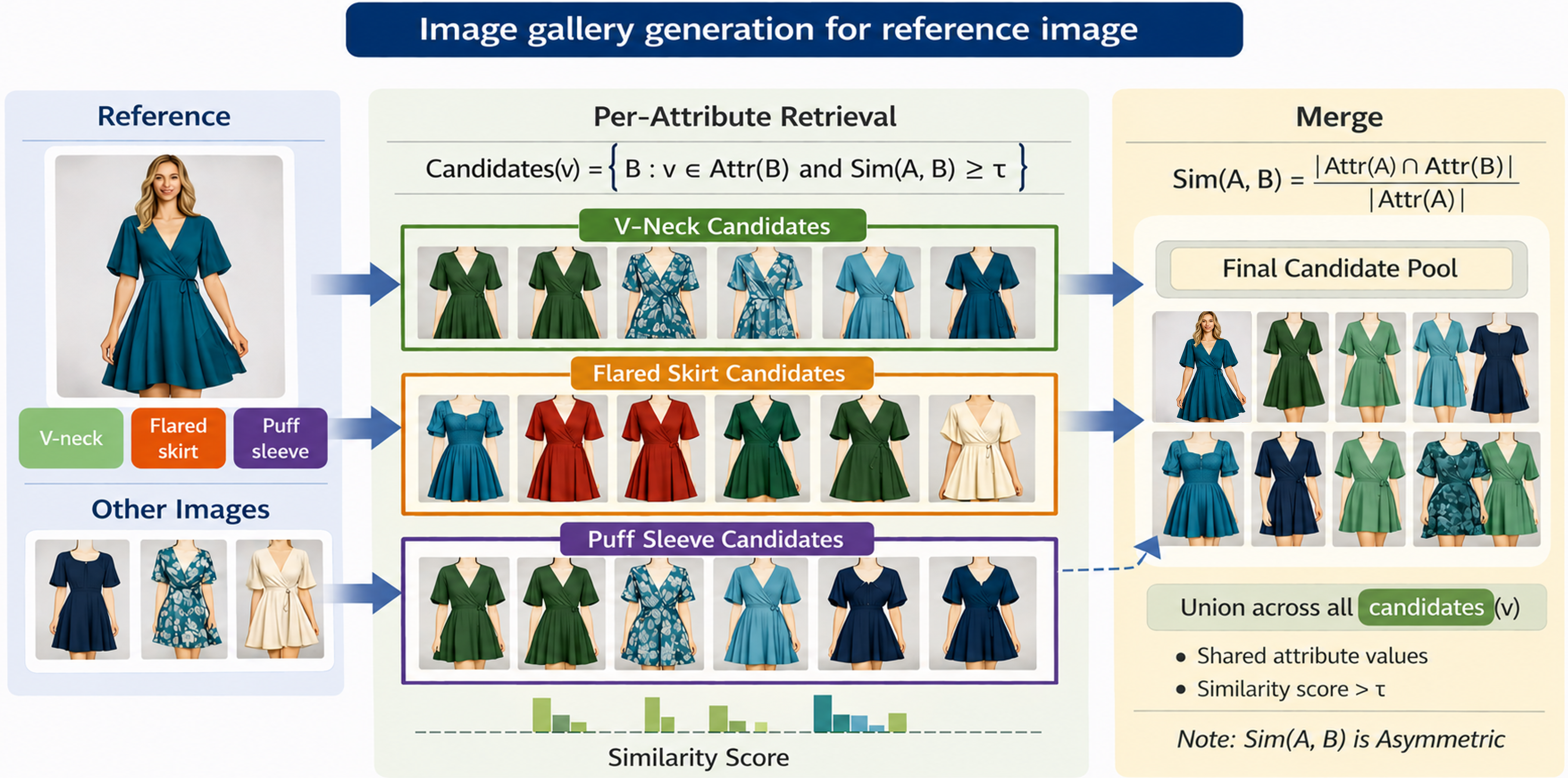}
  \caption{Gallery construction pipeline that retrieves attribute-similar distractors, applies threshold-based difficulty control, and forms hidden-target candidate sets.}
  \label{fig:image_gallery_generation}
\end{figure}

\paragraph{Gallery generation and difficulty control.}
For each target image $A$, we construct a distractor pool by iterating over its
attribute values. For each attribute value, we retrieve a fixed number of images
that share that value with $A$ and satisfy $\mathrm{Sim}(A,B)\ge \tau$, where $\tau$
is a similarity threshold. We then merge the retrieved sets across all attribute
values to obtain the final candidate distractors for $A$. We keep only targets
for which this merged pool contains more than five candidates, and combine each target with a subset of its candidate distractors to
create a gallery. We control difficulty via (i) the threshold $\tau$ (higher
$\tau$ yields more visually similar distractors) and (ii) the gallery size (larger
galleries expand the search space).
Figure~\ref{fig:image_gallery_generation} illustrates this gallery-construction process.

\subsection{Sample image galleries}
\begin{figure}[htbp]
  \centering
  \begin{subfigure}{\textwidth}
    \centering
    \includegraphics[width=0.8\textwidth]{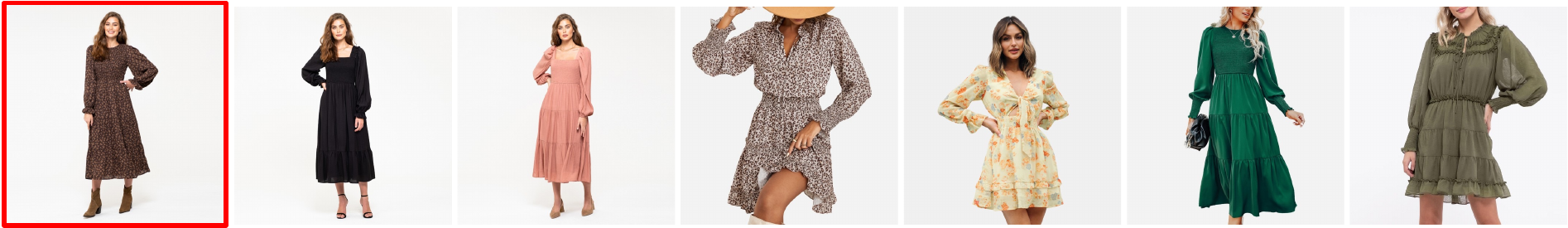}
    \caption{Image Gallery 1}
    \label{fig:gallery1}
  \end{subfigure}

  \begin{subfigure}{\textwidth}
    \centering
    \includegraphics[width=0.8\textwidth]{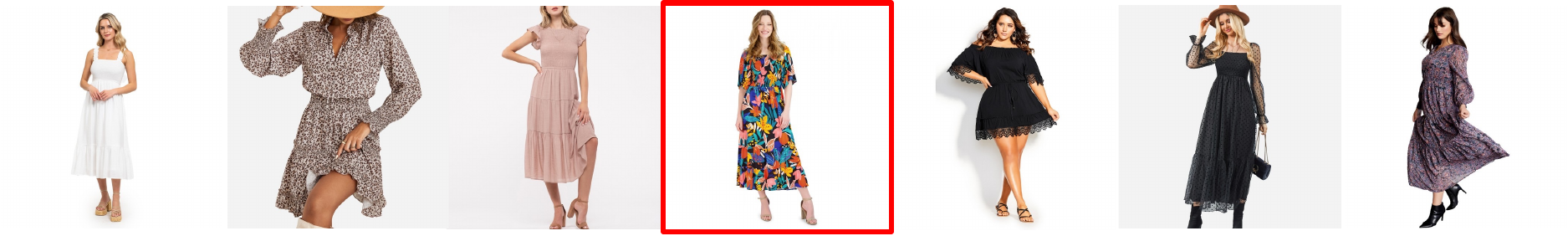}
    \caption{Image Gallery 2}
    \label{fig:gallery2}
  \end{subfigure}

  \begin{subfigure}{\textwidth}
    \centering
    \includegraphics[width=0.8\textwidth]{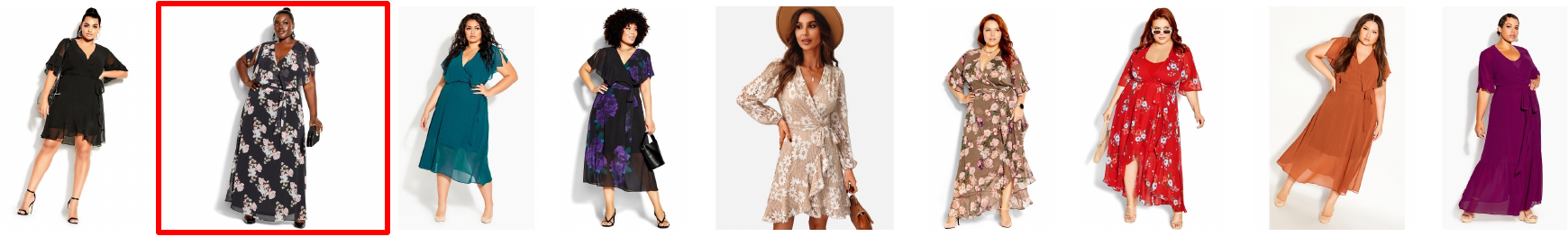}
    \caption{Image Gallery 3}
    \label{fig:gallery3}
  \end{subfigure}

  \begin{subfigure}{\textwidth}
    \centering
    \includegraphics[width=0.8\textwidth]{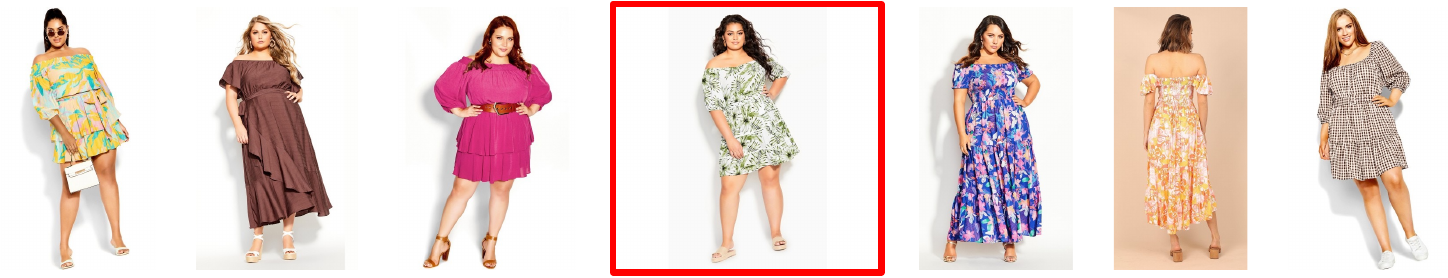}
    \caption{Image Gallery 4}
    \label{fig:gallery4}
  \end{subfigure}

  \caption{Representative \textit{Guess My Preferred Dress} galleries with red-outlined hidden targets.}
  \label{fig:image_galleries}
\end{figure}

Figure~\ref{fig:image_galleries} shows four representative galleries from the dress task, with the individual examples shown in Figures~\ref{fig:gallery1}, \ref{fig:gallery2}, \ref{fig:gallery3}, and~\ref{fig:gallery4}. These galleries illustrate the intended ambiguity of the task: candidates share coarse visual cues such as silhouette and overall style, so useful questions must target subtler construction details rather than prohibited shortcuts.

\paragraph{Scale and release.}
We collect 4{,}880 unique dress images and use five similarity thresholds, $\tau \in \{0.3, 0.4, 0.5, 0.6, 0.8\}$, to control gallery difficulty. Raising $\tau$ keeps only distractors with stronger attribute overlap, so the candidate pool generally contracts while visual confusability increases; lower thresholds admit larger, more varied galleries. We obtain 587 galleries at $\tau=0.8$ because this strict threshold makes it difficult to find enough highly similar distractors, and we release 1{,}000 examples for each of $\tau=0.3$ through $\tau=0.6$, yielding 4{,}587 total benchmark examples in \texttt{anc/data.json} under the CC-BY-4.0 license. Each example contains an ID, a target image URL, and a candidate gallery that includes the target and its similar distractors.

\begin{figure}[htbp]
  \centering
  \begin{subfigure}{0.48\textwidth}
    \centering
    \includegraphics[width=\textwidth]{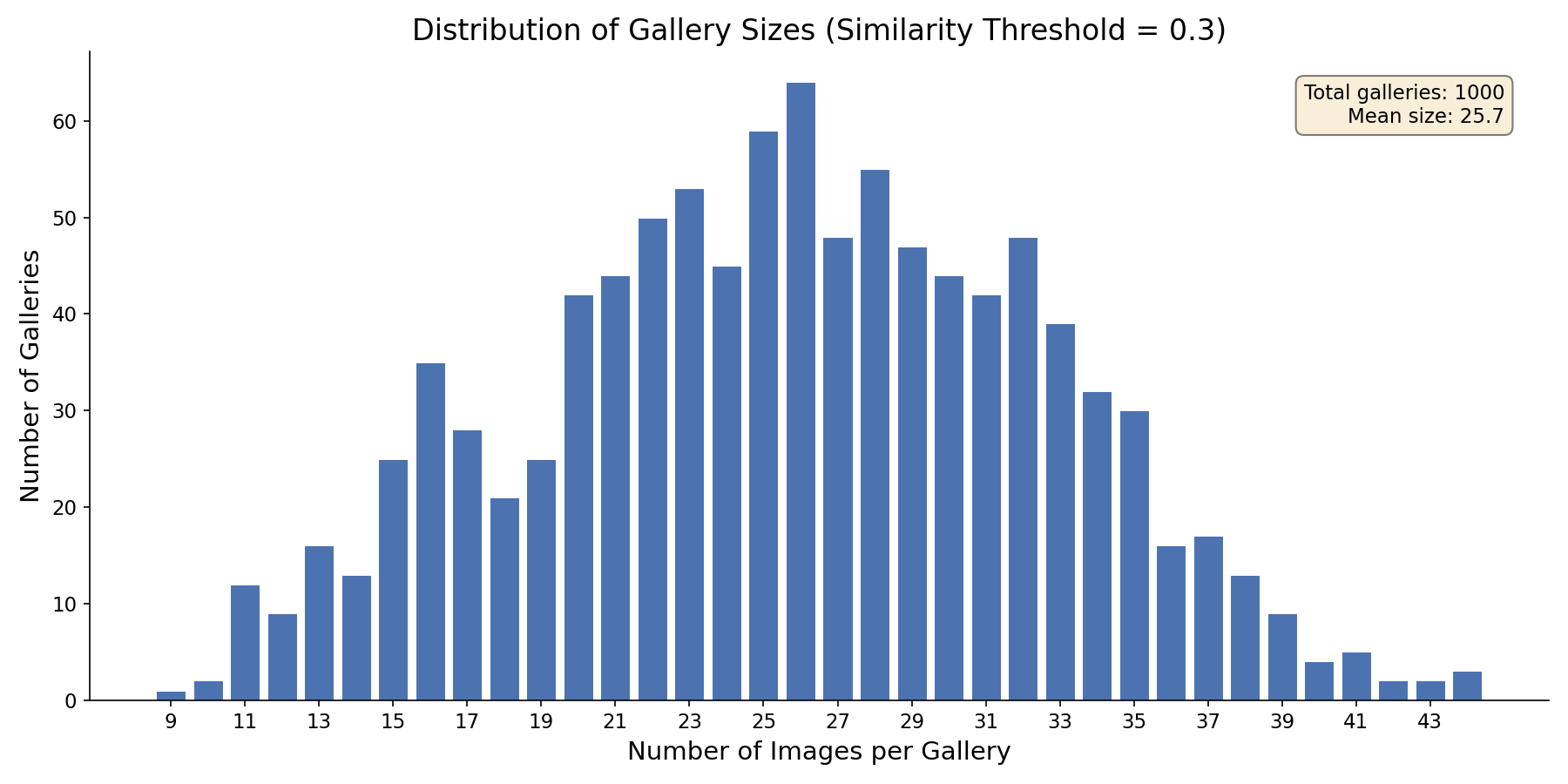}
    \caption{$\tau = 0.3$}
    \label{fig:gsd_03}
  \end{subfigure}
  \hfill
  \begin{subfigure}{0.48\textwidth}
    \centering
    \includegraphics[width=\textwidth]{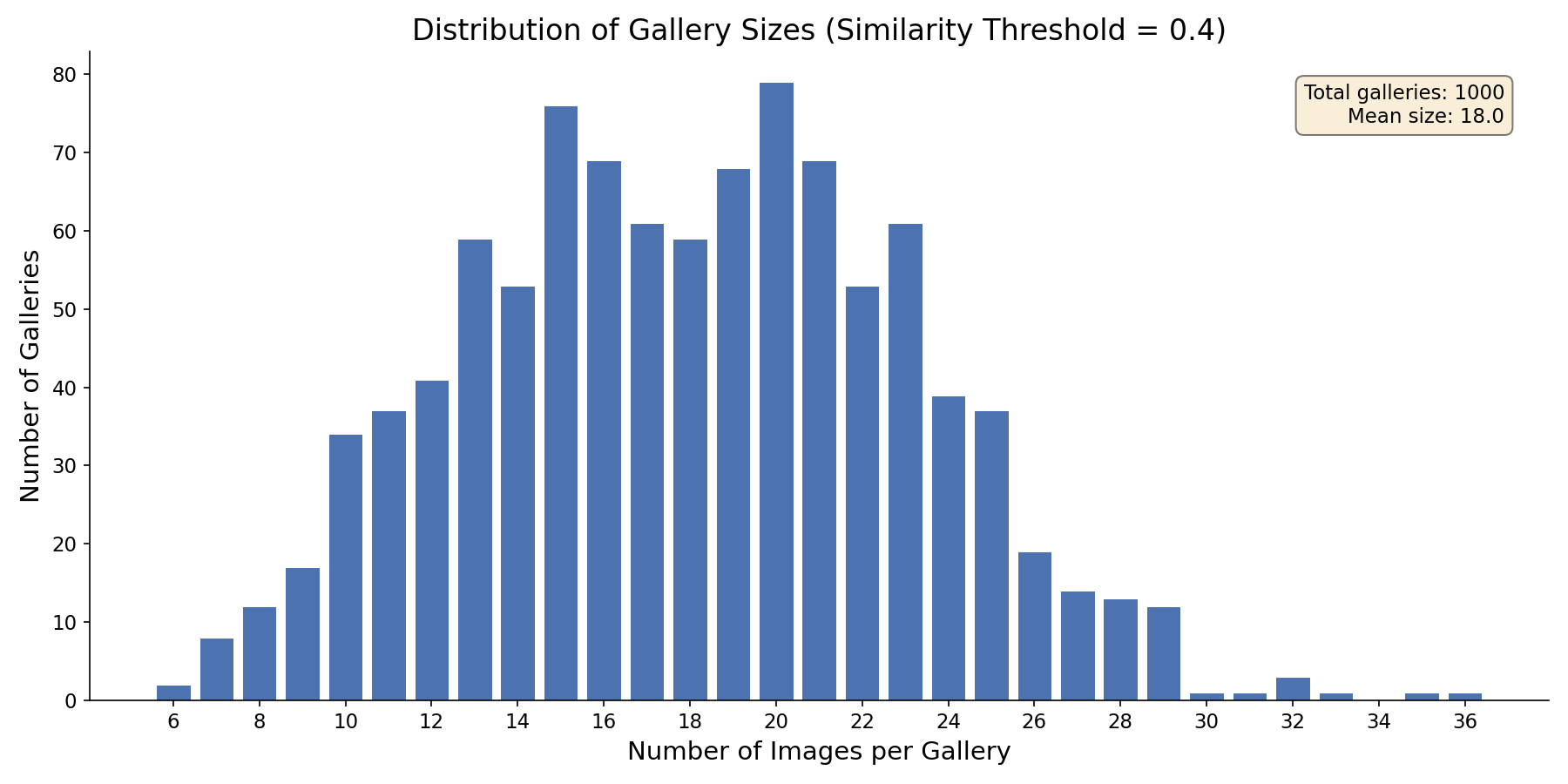}
    \caption{$\tau = 0.4$}
    \label{fig:gsd_04}
  \end{subfigure}

  \vspace{0.5em}
  \begin{subfigure}{0.48\textwidth}
    \centering
    \includegraphics[width=\textwidth]{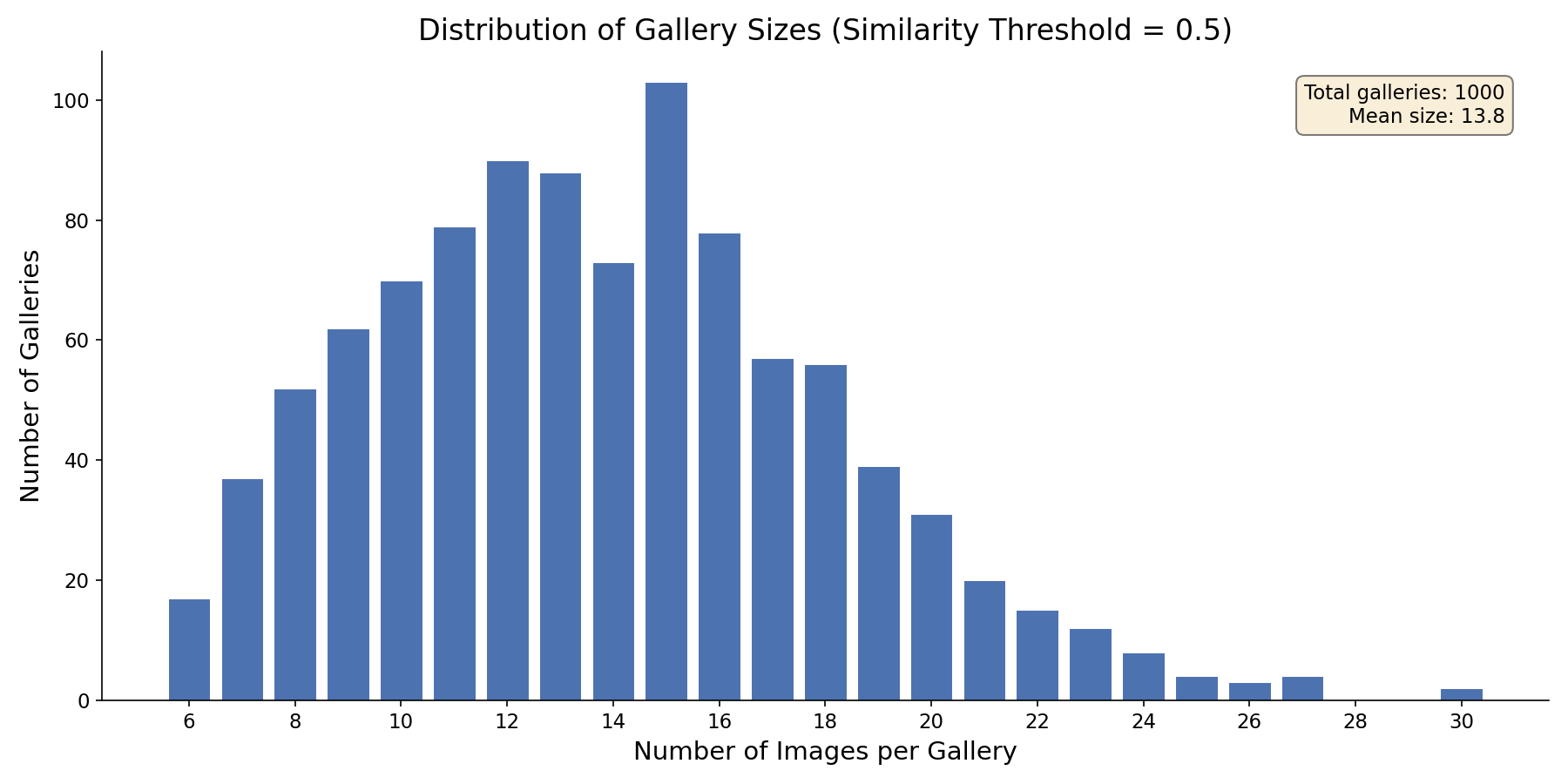}
    \caption{$\tau = 0.5$}
    \label{fig:gsd_05}
  \end{subfigure}
  \hfill
  \begin{subfigure}{0.48\textwidth}
    \centering
    \includegraphics[width=\textwidth]{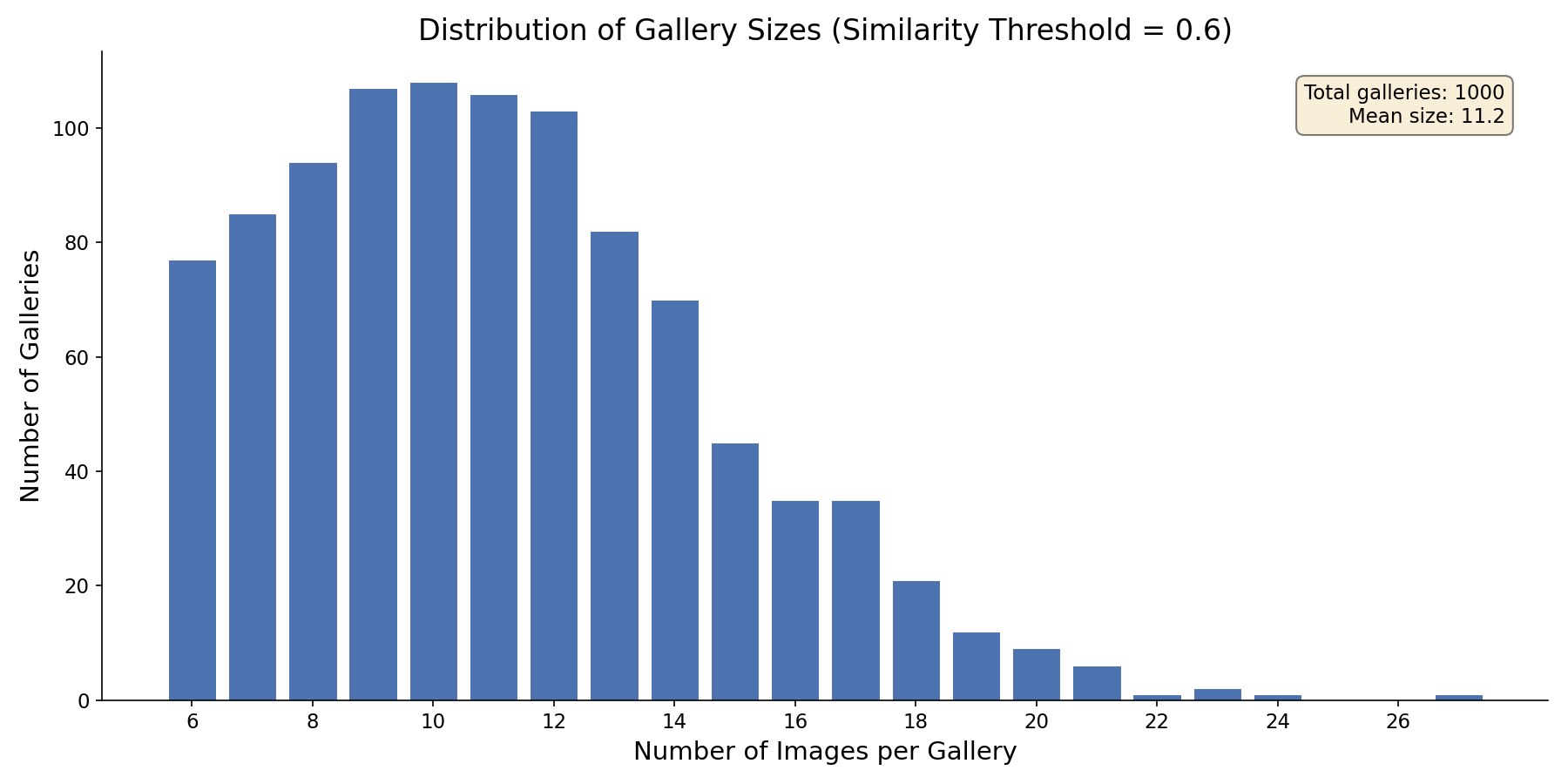}
    \caption{$\tau = 0.6$}
    \label{fig:gsd_06}
  \end{subfigure}

  \vspace{0.5em}
  \begin{subfigure}{0.48\textwidth}
    \centering
    \includegraphics[width=\textwidth]{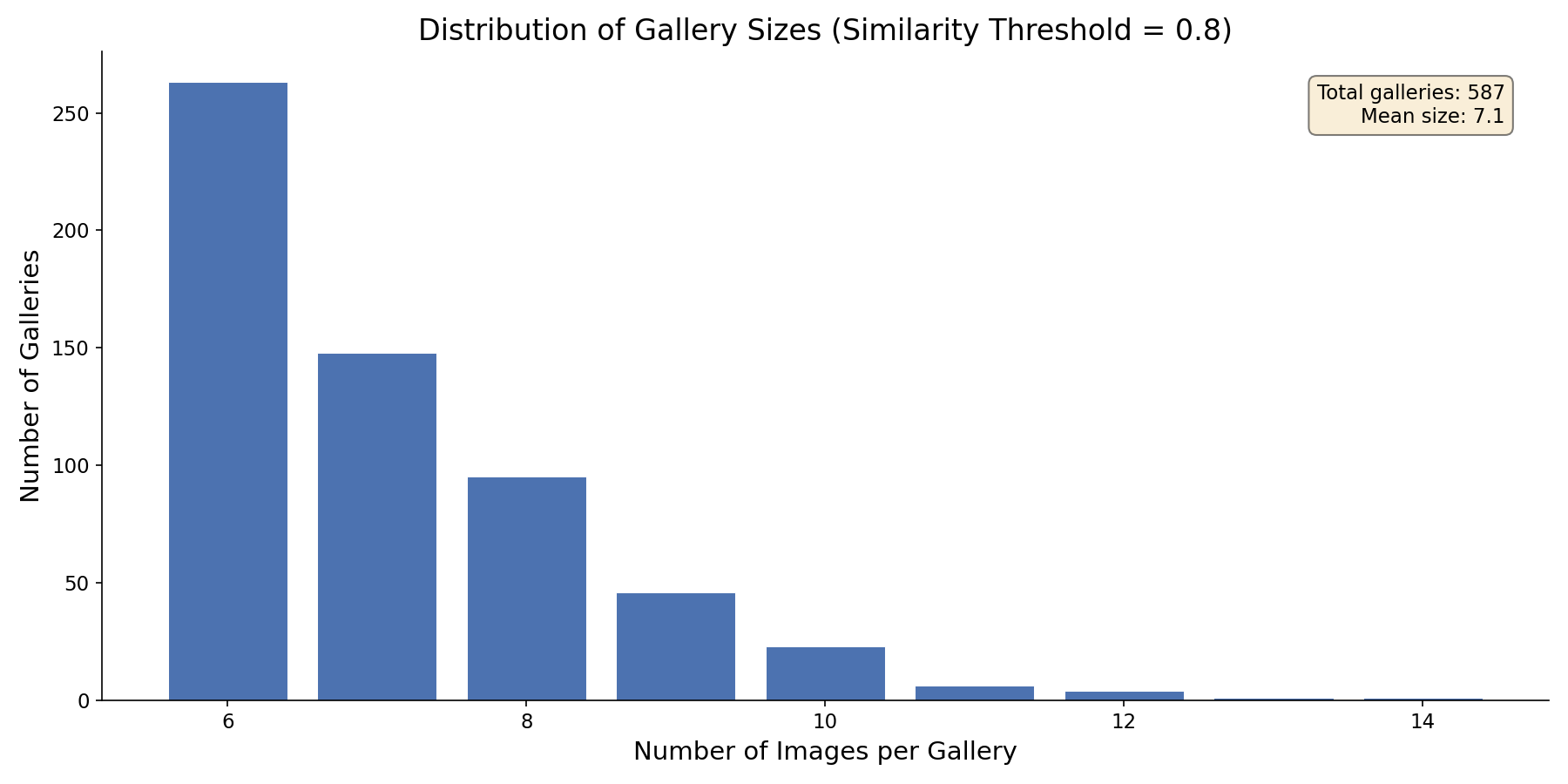}
    \caption{$\tau = 0.8$}
    \label{fig:gsd_08}
  \end{subfigure}

  \caption{Gallery-size distributions under the five similarity thresholds used for difficulty control.}
  \label{fig:gallery_size_dist}
\end{figure}

Table~\ref{tab:released_benchmark_examples} summarizes released and evaluated counts, mean gallery sizes, and random-chance baselines (the expected accuracy of uniformly guessing the target index) by threshold. Due to inference cost, we evaluate a fixed 770-example subset per threshold from $\tau=0.3$ to $\tau=0.6$ and all 587 available examples at $\tau=0.8$. The random-chance baseline, computed as $\mathrm{mean}(1/N_i)$ over the evaluated galleries, rises from 4.2\% at $\tau=0.3$ to 14.5\% at $\tau=0.8$ as galleries become smaller. Figure~\ref{fig:gallery_size_dist} visualizes the gallery-size distributions, with Figures~\ref{fig:gsd_03}, \ref{fig:gsd_04}, \ref{fig:gsd_05}, \ref{fig:gsd_06}, and~\ref{fig:gsd_08} corresponding to $\tau=0.3$, 0.4, 0.5, 0.6, and 0.8, respectively.

\begin{center}
  \small
  \setlength{\tabcolsep}{2.5pt}
  \begin{tabular}{@{}lrrrrrr@{}}
    \toprule
    $\tau$ & 0.3 & 0.4 & 0.5 & 0.6 & 0.8 & Total \\
    \midrule
    Released & 1{,}000 & 1{,}000 & 1{,}000 & 1{,}000 & 587 & 4{,}587 \\
    Evaluated & 770 & 770 & 770 & 770 & 587 & 3{,}667 \\
    Mean gallery size & 25.7 & 18.0 & 13.8 & 11.2 & 7.1 & -- \\
    Random chance (\%) & 4.2 & 6.1 & 7.9 & 9.9 & 14.5 & -- \\
    \bottomrule
  \end{tabular}
  {\captionsetup{hypcap=false}
  \captionof{table}{Released and evaluated examples, mean gallery sizes, and random-chance baselines ($\mathrm{mean}(1/N_i)$ over evaluated examples) by similarity threshold. Gallery size includes the hidden target.}
  \label{tab:released_benchmark_examples}}
\end{center}

\section{Benchmark framework}
\begin{figure}[htbp]
  \centering
  \includegraphics[width=\textwidth]{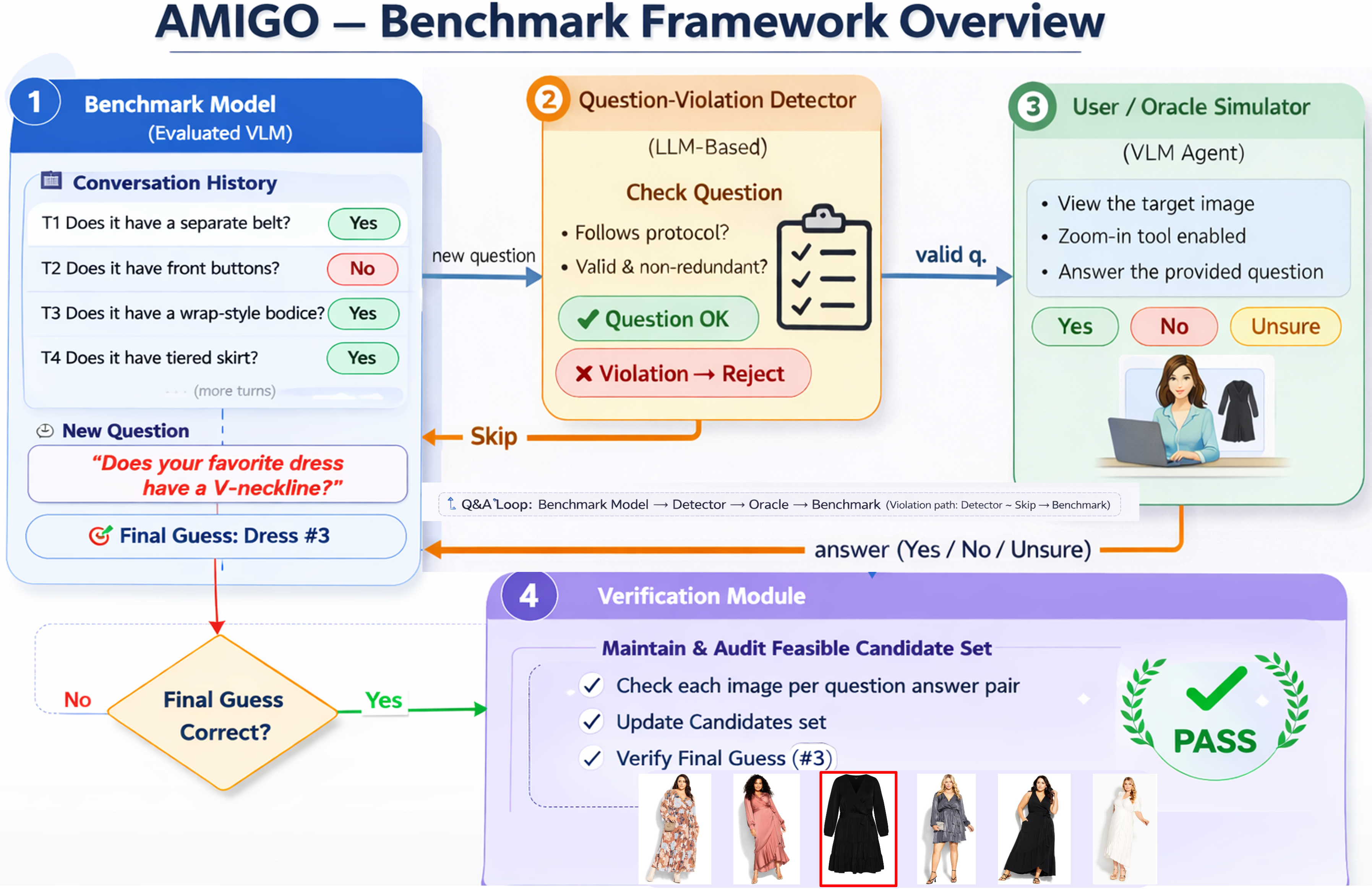}
  \caption{AMIGO evaluation framework linking model questions, protocol checking, oracle feedback, and evidence verification.}
  \label{fig:benchmark_framework_overview}
\end{figure}

\paragraph{Benchmark components.}
Our framework comprises four modules (Figure~\ref{fig:benchmark_framework_overview}):
(i) a \textbf{benchmark model} (the evaluated VLM) that generates questions and produces a final guess,
(ii) a \textbf{question-violation detector} (LLM-based) that enforces the protocol, flags invalid questions with \emph{Skip}, and provides standardized feedback on protocol adherence,
(iii) a \textbf{user/oracle simulator} (VLM agent) that answers valid questions with Yes/No/Unsure, and
(iv) a \textbf{verification module} that maintains and audits the feasible candidate set implied by the dialogue.

Given the uploaded gallery and dialogue history, the benchmark model outputs either the next question or a terminal guess. 

To make question-violation detector and oracle feedback more robust, each newly generated question is paraphrased into two additional variants. Feedback from both the question-violation detector and the oracle is then determined by majority vote across the three question variants.

The violation detector checks each question against the benchmark constraints. It first applies deterministic P1 format checks for empty turns, non-question responses, multi-question turns, and questions that cannot be answered with Yes/No/Unsure. It then applies two few-shot LLM-based checks: one flags prohibited-attribute questions and image-index references, and the other flags repeated enumeration of an attribute type that has already appeared in previous valid turns. The prompts for these two checks are provided in the Appendix (Figures~\ref{fig:question_type_verification_prompt} and~\ref{fig:detect_attribute_from_history_prompt}). Any violation returns \emph{Skip} with no oracle information. We record the number (and rate) of \emph{Skip} responses per trajectory as the primary signal of protocol compliance. Implementing the violation detector as a separate module ensures that all models receive consistent, standardized feedback on protocol adherence regardless of their internal architecture or reasoning style.

The user/oracle simulator answers \emph{valid} questions with \emph{Yes}/\emph{No}/\emph{Unsure}.

\paragraph{Verification module (state tracking and consistency checking).}
This separate evaluation module verifies whether a correct final guess is supported by a coherent process that systematically narrows and ranks the candidate space. The verification module is intentionally \emph{model-relative}: it uses the same evaluated benchmark model to obtain candidate-level responses so that verification reflects the model's own visual judgments and reasoning logic, rather than those of an external verifier. Accordingly, a verification pass measures \emph{model-relative evidence consistency}: whether the evaluated model's candidate-level judgments support the target as the unique best candidate under the accumulated dialogue evidence. It should not be interpreted as model-independent proof of visual correctness.

After each \emph{valid} question--answer pair, the verification module converts the interaction into an explicit constraint and applies it to the gallery to update the \emph{feasible candidate set}, defined as the subset of images that remain consistent with all constraints observed so far. Let $\mathcal{G}$ denote the full gallery of candidate images, and let $\mathcal{F}_t \subseteq \mathcal{G}$ denote the feasible candidate set after the $t$-th valid question--answer pair, with $\mathcal{F}_0=\mathcal{G}$.

To mimic the benchmarked model's reasoning process, the verification module replays the three question variants
$\mathcal{Q}_t=\{q_t^{(1)},q_t^{(2)},q_t^{(3)}\}$
on each candidate image $x_i \in \mathcal{F}_{t-1}$ and obtains the corresponding responses from the same benchmark model being evaluated
$\mathcal{R}_{i,t}=\{r_{i,t}^{(1)},r_{i,t}^{(2)},r_{i,t}^{(3)}\}$.
The dominant response is determined by majority voting:
\begin{equation}
\hat{r}_{i,t}
=
\arg\max_{a \in \mathcal{A}}
\sum_{k=1}^{3}
\mathbf{1}\!\left[r_{i,t}^{(k)}=a\right].
\end{equation}

where $\mathcal{A}$ denotes the answer space. We define the confidence score of the dominant response as its vote share:
\begin{equation} 
c_{i,t} = \begin{cases} 0.5, & \text{if } \texttt{Unsure} \in \mathcal{R}_{i,t},\\[2mm] \frac{1}{3} \max\limits_{a \in \mathcal{A}} \sum\limits_{k=1}^{3} \mathbf{1}\!\left[r_{i,t}^{(k)}=a\right], & \text{otherwise}
\end{cases} 
\end{equation}

Thus, $c_{i,t}=1$ when all three variants produce the same response, $c_{i,t}=2/3$ when two variants agree and one differs, and $c_{i,t}=0.5$ whenever \texttt{Unsure} appears in the response set.

A candidate image is eliminated only when its dominant response is fully confident but inconsistent with the oracle's aggregated answer $a_t^{\mathrm{orc}}$ from the $t$-th question--answer pair. Accordingly, the feasible candidate set is updated as
\begin{equation}
\mathcal{F}_t = \left\{ x_i \in \mathcal{F}_{t-1} : c_{i,t}<1 \;\;\text{or}\;\; \hat{r}_{i,t}=a_t^{\mathrm{orc}} \right\}
\end{equation}

For each remaining candidate image, the module maintains an accumulated alignment score with respect to the oracle's answers. Let $S_{i,t}$ denote the accumulated alignment score of image $x_i$ after turn $t$, initialized as $S_{i,0}=0$. The score is updated as
\begin{equation}
S_{i,t}
=
S_{i,t-1}
+
\mathbf{1}\!\left[\hat{r}_{i,t}=a_t^{\mathrm{orc}}\right]c_{i,t}
+
\mathbf{1}\!\left[\hat{r}_{i,t}\neq a_t^{\mathrm{orc}}\right](1-c_{i,t})
\end{equation}
This update rewards candidates whose dominant response agrees with the oracle in proportion to their confidence, while assigning a complementary score when the dominant response disagrees with the oracle.

If the feasible candidate set is reduced to either a singleton or an empty set before the final question--answer pair is reached, the module reverts to the previous feasible candidate set and stops applying further hard elimination. It then applies the remaining questions to all images in the reverted candidate set and continues updating only their alignment scores. This rollback prevents an intermediate singleton from determining the verification result before all valid dialogue evidence has been incorporated. Let $\mathcal{F}^{\mathrm{final}}$ denote the final active candidate set after any such rollback. The set of top-scoring remaining candidates is
\begin{equation}
\mathcal{M}_T
=
\operatorname*{arg\,max}_{x_i \in \mathcal{F}^{\mathrm{final}}}
S_{i,T},
\end{equation}
where $T$ is the total number of valid question--answer pairs. Let $x^*$ denote the hidden target. A trajectory passes verification if and only if $\mathcal{M}_T=\{x^*\}$; that is, the target is the unique highest-scoring candidate. If multiple candidates tie for the highest alignment score, the trajectory does not pass verification.

This module serves three purposes:
(1) \textbf{Evidence verification:} determining whether the dialogue makes the target the unique highest-scoring candidate under the evaluated model's candidate-level judgments;
(2) \textbf{Trajectory auditing:} logging candidate-set reduction dynamics, such as elimination rates and periods of stagnation, to support trajectory-level diagnostics; and
(3) \textbf{Consistency checking:} detecting contradictions such as an empty feasible candidate set or conflicting constraints, thereby supporting trajectory-level analysis of whether the benchmark model recovers conservatively rather than guessing prematurely.

Importantly, the feasible candidate set is updated only after valid, non-\emph{Skip} question--answer pairs that eliminate at least one candidate. This design cleanly separates protocol violations from evidence-based candidate elimination.

\section{Evaluation Metrics}
We evaluate models along four complementary axes. Unless noted otherwise, all constraint application and candidate-set updates are computed by the \textbf{verification module} using only \emph{valid} (non-\texttt{Skip}) question-answer pairs.
\begin{itemize}
	  \item \textbf{Identification accuracy: Unverified Overall, Verified Correct, Random-Guess Correct.}
	 \emph{Unverified Overall accuracy} measures whether the model's final guess matches the hidden target.
	  Trajectories that exhaust the 20-question budget without a final guess are treated as incorrect terminal outcomes.
	  Here, \emph{Unverified Overall} refers to this final-guess correctness rate and is partitioned into two mutually exclusive correct-outcome categories: \emph{Unverified Overall} = \emph{Verified Correct} + \emph{Random-Guess Correct}.
	  \emph{Verified Correct accuracy} excludes random-but-correct guesses: a trajectory is counted as verified-successful only if
  (i) the final guess is correct and 
  (ii) the verification module produces the hidden target as the unique highest-scoring candidate. This metric captures whether the model's own candidate-level judgments support its correct final answer with a unique top-ranked target.
  \emph{Random-Guess Correct accuracy} counts trajectories where the final guess is correct but the trajectory fails the verification module because the target is not the unique highest-scoring remaining candidate. This category captures correctness without sufficient model-relative, verification-passing evidence.
  \item \textbf{Interaction efficiency.}
  We measure interaction cost as the number of total turns (including \textit{Skip}) after the \textit{End of uploading} signal and before the final guess.
  We report efficiency separately for Verified Correct, Random-Guess Correct, incorrect, and overall trajectories to distinguish evidence-driven efficiency from premature or unproductive stopping.
  \item \textbf{Protocol compliance.}
  We quantify instruction following with (i) the average \textit{Skip} rate and (ii) the premature-output rate before the \textit{End of uploading} signal. Non-question acknowledgments during the upload phase are not counted as premature outputs.
  Lower average \textit{Skip} rates and zero premature outputs reflect stronger adherence; repeated violations reflect failures to recover to valid questioning.
  \item \textbf{Robustness under controlled noisy feedback.}
  Recovery rate measures the fraction of perturbed successful trajectories in which the model still produces the correct final guess. It is conditional on clean verified success, not all attempted dialogues.
\end{itemize}

\section{Experimental results}

\definecolor{amigoQwen3}{HTML}{4C72B0}
\definecolor{amigoQwen35}{HTML}{DD8452}
\definecolor{amigoStep3}{HTML}{55A868}
\definecolor{amigoGemma31}{HTML}{C44E52}
\definecolor{amigoGemma12}{HTML}{8172B3}
\newcommand{\legenditem}[2]{\raisebox{0.15ex}{\textcolor{#1}{\rule{0.85em}{0.85em}}}\hspace{0.3em}#2}
\newcommand{\modelcolorlegend}{%
  \par\vspace{2pt}
  {\footnotesize
  \begin{tabular}{@{}lll@{}}
    \legenditem{amigoQwen3}{Qwen3-VL} &
    \legenditem{amigoQwen35}{Qwen3.5-397B} &
    \legenditem{amigoStep3}{Step3-VL-10B} \\
    \legenditem{amigoGemma31}{Gemma4-31B} &
    \legenditem{amigoGemma12}{Gemma4-12B} &
  \end{tabular}}
}

We evaluate five open-source VLMs on \textit{Guess My Preferred Dress} across similarity thresholds $\tau \in \{0.3,0.4,0.5,0.6,0.8\}$:
\textbf{Qwen3-VL-235B-A22B-Instruct-FP8}~\cite{qwen3},
\textbf{Qwen3.5-397B-A17B-FP8}~\cite{qwen35},
\textbf{Step3-VL-10B}~\cite{step3vl10b},
\textbf{Gemma4-31B}~\cite{gemma4-31b}, and
\textbf{Gemma4-12B}~\cite{gemma4-12b}. We evaluate 770 samples per model and threshold from $\tau=0.3$ to $\tau=0.6$ (out of 1{,}000 released examples per threshold), and all 587 available samples at $\tau=0.8$. Mean gallery sizes decrease from 25.7 images at $\tau=0.3$ to 7.1 at $\tau=0.8$ (Table~\ref{tab:released_benchmark_examples}). At each threshold, all models are evaluated on the same fixed examples, so cross-model comparisons are paired by gallery.

\paragraph{Evaluation setup.}
We use \textbf{GLM-5.2}~\cite{glm52} as the question-violation detector. The oracle uses an agentic Qwen-Agent pipeline with an image zoom-in tool and \textbf{Qwen3-VL-235B-A22B-Instruct-FP8}~\cite{qwen3} as the VLM backbone. For the main run, configurable thinking is disabled: Qwen-family pilot runs produced non-terminating or malformed reasoning traces even with a 16{,}384-token budget, while Gemma4 thinking was prohibitively slow; \textbf{Step3-VL-10B} keeps its default thinking behavior. The same inference setting is used when a benchmarked model is called inside the verification module. Because Step3-VL-10B retains thinking while all other models have it disabled, cross-model comparisons should be interpreted as comparisons under a practical inference configuration rather than a fully controlled architecture comparison. We also report $\tau=0.8$-only results across all metric axes for \textbf{Gemma4-31B} with thinking enabled and \textbf{Muse-Glimmer-30B}~\cite{museglimmer} with default reasoning.

\subsection{Oracle reliability checks}
The oracle is a core measurement component: it turns each valid natural-language question into a \emph{Yes}/\emph{No}/\emph{Unsure} signal about the hidden target image. We therefore test whether the reported model differences are likely to be driven by the benchmarked models rather than by unstable oracle feedback. The audit uses two complementary checks. The cross-backbone check measures sensitivity to the VLM used inside the oracle, while the human expert check measures agreement with an external fashion-domain judge.

Both checks show high agreement with the original oracle. The cross-backbone audit obtains 90.8\% agreement on 3,327 valid target-image/question pairs ($\kappa=0.816$; Table~\ref{tab:oracle_reliability}). The human fashion-expert audit obtains 95.2\% agreement on 500 valid pairs ($\kappa=0.90$; Table~\ref{tab:human_oracle_audit}). We report the per-model breakdowns below to verify that reliability is not concentrated in only one model's trajectories.

\paragraph{Cross-backbone consistency.}
For the cross-backbone check, we rerun the same Qwen-Agent oracle pipeline with a different VLM backbone, \textbf{Qwen3.5-397B-A17B-FP8}. For each benchmarked model and each similarity threshold, we randomly sample 20 interaction trajectories and extract all valid target-image/question pairs. The Qwen3.5-backed oracle answers each pair independently, and we compare its response with the original Qwen3-VL-backed oracle response used in the benchmark.

The cross-backbone audit is stratified by the benchmarked model whose trajectories supplied the sampled pairs. Table~\ref{tab:oracle_reliability} reports the per-model pair counts, agreement, and Cohen's $\kappa$, which accounts for chance agreement under the observed response distribution.

The cross-backbone disagreements show a mild asymmetry: the Qwen3.5-backed oracle answers ``Yes'' in 37.7\% of cases, compared with 43.2\% for the original oracle. This difference is not concentrated on a particular benchmarked model or similarity threshold, suggesting that residual disagreements mainly reflect ambiguous borderline questions rather than systematic model- or threshold-specific bias.

\begin{table}[htbp]
  \centering
  \small
  \setlength{\tabcolsep}{4pt}
  \resizebox{\textwidth}{!}{%
  \begin{tabular}{@{}lccc@{}}
    \toprule
    \textbf{Model} & \textbf{\# Pairs} & \textbf{Agreement (\%)} & \textbf{Cohen's $\kappa$} \\
    \midrule
    \textbf{Gemma4-31B} & 515 & 92.2 & 0.844 \\
    \textbf{Gemma4-12B} & 913 & 89.7 & 0.789 \\
    \textbf{Qwen3.5-397B-A17B-FP8} & 811 & 91.5 & 0.820 \\
    \textbf{Qwen3-VL-235B-A22B-Instruct-FP8} & 487 & 92.4 & 0.846 \\
    \textbf{Step3-VL-10B} & 601 & 90.7 & 0.796 \\
    \midrule
    \textbf{Overall} & \textbf{3,327} & \textbf{90.8} & \textbf{0.816} \\
    \bottomrule
  \end{tabular}%
  }
  \caption{Cross-backbone oracle reliability audit across sampled valid question pairs.}
  \label{tab:oracle_reliability}
\end{table}

\noindent
Numerically, the cross-backbone results are tightly clustered across source trajectories: per-model agreement ranges from 89.7\% to 92.4\%, and Cohen's $\kappa$ ranges from 0.789 to 0.846. The lowest-agreement row, \textbf{Gemma4-12B}, still remains near 90\% agreement despite contributing the largest sampled set (913 pairs), while the highest-agreement row, \textbf{Qwen3-VL-235B-A22B-Instruct-FP8}, reaches 92.4\%. The 2.7-point agreement spread and 0.057 $\kappa$ spread indicate that oracle consistency is not concentrated in one benchmarked model's trajectories. The remaining 9.2\% cross-backbone disagreement rate is therefore best interpreted as residual ambiguity in natural-language fashion questions rather than a model-specific oracle failure.

\paragraph{Human expert agreement.}
For the human audit, we randomly sample 100 valid target-image/question pairs from each benchmarked model's trajectories. A fashion expert knowledgeable about garment terminology answers each question by inspecting the corresponding target image, and we compare the expert answer with the original oracle answer used in the benchmark. Table~\ref{tab:human_oracle_audit} reports the per-model human-audit agreement and Cohen's $\kappa$ values.

\begin{table}[htbp]
  \centering
  \small
  \setlength{\tabcolsep}{4pt}
  \resizebox{\textwidth}{!}{%
  \begin{tabular}{@{}lccc@{}}
    \toprule
    \textbf{Model} & \textbf{\# Pairs} & \textbf{Agreement (\%)} & \textbf{Cohen's $\kappa$} \\
    \midrule
    \textbf{Gemma4-31B} & 100 & 93.0 & 0.859 \\
    \textbf{Gemma4-12B} & 100 & 96.0 & 0.920 \\
    \textbf{Qwen3.5-397B-A17B-FP8} & 100 & 97.0 & 0.940 \\
    \textbf{Qwen3-VL-235B-A22B-Instruct-FP8} & 100 & 95.0 & 0.900 \\
    \textbf{Step3-VL-10B} & 100 & 95.0 & 0.880 \\
        \midrule
    \textbf{Overall} & \textbf{500} & \textbf{95.2} & \textbf{0.90} \\
    \bottomrule
  \end{tabular}%
  }
  \caption{Human fashion-expert oracle reliability audit across sampled valid question pairs.}
  \label{tab:human_oracle_audit}
\end{table}

\noindent
The human audit provides an even stronger reliability signal: agreement ranges from 93.0\% to 97.0\%, $\kappa$ ranges from 0.859 to 0.940, and only 24 of 500 audited pairs disagree with the original oracle. Even the weakest row, \textbf{Gemma4-31B}, has 93.0\% agreement and $\kappa=0.859$, while \textbf{Qwen3.5-397B-A17B-FP8} reaches 97.0\% agreement and $\kappa=0.940$. The overall human agreement is 4.4 percentage points higher than the cross-backbone agreement (95.2\% vs. 90.8\%), suggesting that the larger cross-backbone discrepancy mainly reflects differences between automated VLM backbones on borderline wording or visual evidence, not a systematic mismatch between the oracle and human fashion judgment.

\subsection{Question-violation detector audit}
We audit mixed non-empty questions using 200 samples per model pooled across thresholds; each sample contains the target image URL, audited question, and prior valid-question history. As shown in Table~\ref{tab:violation_detector_human_audit_main}, detector-human agreement is 94.0\% across 1{,}000 questions ($\kappa=0.868$). Under the audit convention used for the table, treating invalid questions as positive gives 86.1\% precision, 97.2\% recall, and 91.3\% F1.

\begin{table}[!t]
  \centering
  \small
  \setlength{\tabcolsep}{1.5pt}
  \renewcommand{\arraystretch}{0.95}
  \begin{tabular}{@{}lcccc@{}}
    \toprule
    \textbf{Model} & \textbf{Raw agr.} & $\mathbf{P_{inv}}$ & $\mathbf{R_{inv}}$ & $\boldsymbol{\kappa}$ \\
    \midrule
    Qwen3-VL & 95.0 & 80.8 & 100.0 & 0.861 \\
    Qwen3.5-397B & 92.0 & 85.7 & 93.0 & 0.829 \\
    Step3-VL-10B & 97.0 & 94.7 & 90.0 & 0.904 \\
    Gemma4-31B & 95.5 & 90.1 & 100.0 & 0.909 \\
    Gemma4-12B & 90.5 & 82.6 & 100.0 & 0.812 \\
    \midrule
    \textbf{Overall} & \textbf{94.0} & \textbf{86.1} & \textbf{97.2} & \textbf{0.868} \\
    \bottomrule
  \end{tabular}
  \caption{Human audit of the question-violation detector on mixed non-empty questions. Each row uses 200 samples pooled across thresholds; Raw agr., $P_{inv}$, $R_{inv}$, and $\kappa$ denote raw agreement, invalid-class precision, invalid-class recall, and Cohen's kappa under the audit convention used for this table. Percentage columns use \%.}
  \label{tab:violation_detector_human_audit_main}
\end{table}

The per-model detector audit values remain in a high-agreement regime: raw agreement ranges from 90.5\% to 97.0\%, and $\kappa$ ranges from 0.812 to 0.909. The table also shows variation in the invalid-class precision-recall tradeoff: recall is 100.0\% for Qwen3-VL, Gemma4-31B, and Gemma4-12B, 93.0\% for Qwen3.5-397B, and 90.0\% for Step3-VL-10B, while precision ranges from 80.8\% to 94.7\%. The direction of disagreement matters for interpreting the benchmark: in 51 of the 60 mismatches, human annotators label the question invalid but the detector predicts it as valid. This makes the audit bias permissive: some prohibited or repeated questions can receive oracle information instead of \texttt{Skip}, so \texttt{Skip}-based violation rates should be read as conservative estimates of the true invalid-action frequency.

The remainder of this section analyzes the result set along four axes: accuracy and evidence grounding, efficiency, protocol compliance, and robustness.

\subsection{Accuracy and evidence grounding}
\begin{figure*}[!t]
  \centering
  \includegraphics[width=\textwidth]{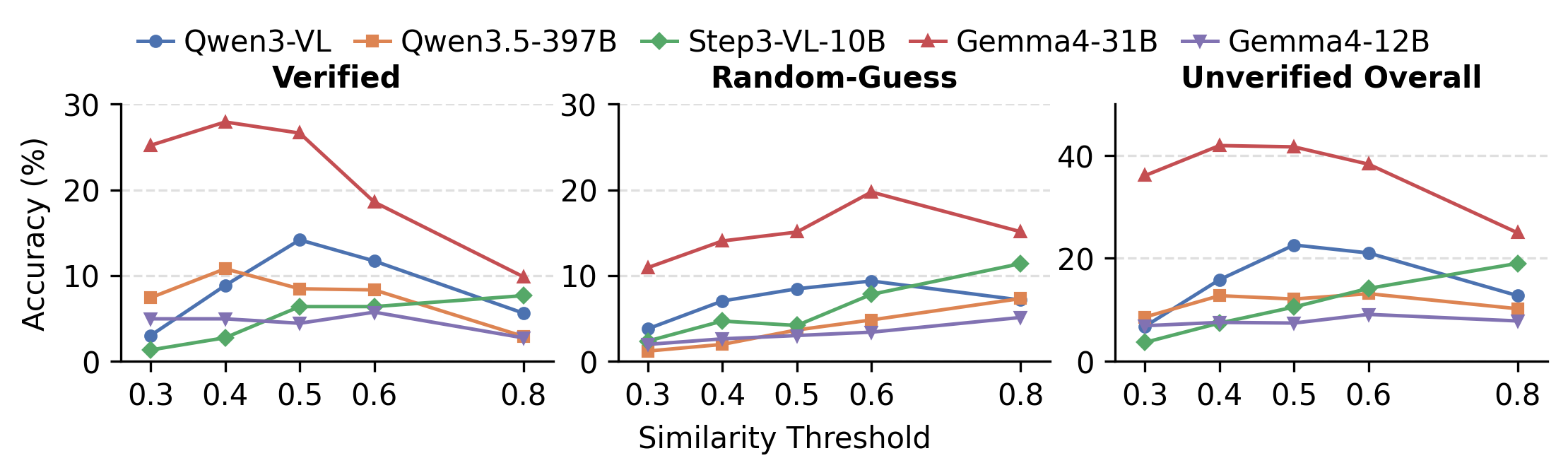}
  \caption{Accuracy by outcome category across similarity thresholds.}
  \label{fig:all_models_accuracies}
\end{figure*}

Figure~\ref{fig:all_models_accuracies} and Table~\ref{tab:per_threshold_accuracies_all_models} report Verified Correct, Random-Guess Correct, and Unverified Overall Accuracy across similarity thresholds. Verified Correct is the main evidence-grounded identification metric, while the gap between Unverified Overall and Verified Correct captures correct final guesses that are not supported by a verification-passing trajectory.

The accuracy profiles differ substantially across models. \textbf{Gemma4-31B} has Verified Correct accuracy above 25\% from $\tau=0.3$ to $\tau=0.5$ before declining at stricter thresholds. \textbf{Qwen3-VL-235B-A22B-Instruct-FP8} has a middle-threshold peak: it rises from 3.0\% Verified at $\tau=0.3$ to 8.8\% at $\tau=0.4$ and 14.2\% at $\tau=0.5$, then remains elevated at 11.7\% at $\tau=0.6$. \textbf{Qwen3.5-397B-A17B-FP8} shows a split accuracy profile: Verified Correct accuracy peaks at $\tau=0.4$ (10.8\%), holds near 8\% through $\tau=0.5$--$0.6$, then drops sharply to 2.9\% at $\tau=0.8$, while Unverified Overall Accuracy remains comparatively stable from $\tau=0.4$ to $\tau=0.6$ (12.1--13.1\%) and still stands at 10.2\% at $\tau=0.8$ because the decline in Verified Correct is partially offset by rising Random-Guess Correct accuracy. \textbf{Step3-VL-10B} improves as galleries become smaller, reaching its best Verified Correct and Unverified Overall Accuracy at $\tau=0.8$. \textbf{Gemma4-12B} reaches 4.9\% Verified Correct accuracy at both $\tau=0.3$ and $\tau=0.4$, but does not gain as galleries become smaller.

These patterns lead to three accuracy-specific observations. First, \textbf{Gemma4-31B} has the clearest evidence-grounded advantage from low-to-middle similarity thresholds. Second, \textbf{Qwen3-VL-235B-A22B-Instruct-FP8} has the strongest non-Gemma4-31B Unverified Overall Accuracy from $\tau=0.4$ to $\tau=0.6$, and its verified advantage emerges most clearly at $\tau=0.5$ and $\tau=0.6$. Third, \textbf{Step3-VL-10B} has the opposite threshold profile, remaining weaker through the middle thresholds but becoming the strongest non-Gemma4-31B model in the highest-$\tau$, smallest-gallery setting, where it reaches 7.6\% Verified and 19.0\% Unverified Overall Accuracy.

\begin{figure*}[!htbp]
  \centering
  \includegraphics[width=0.85\textwidth]{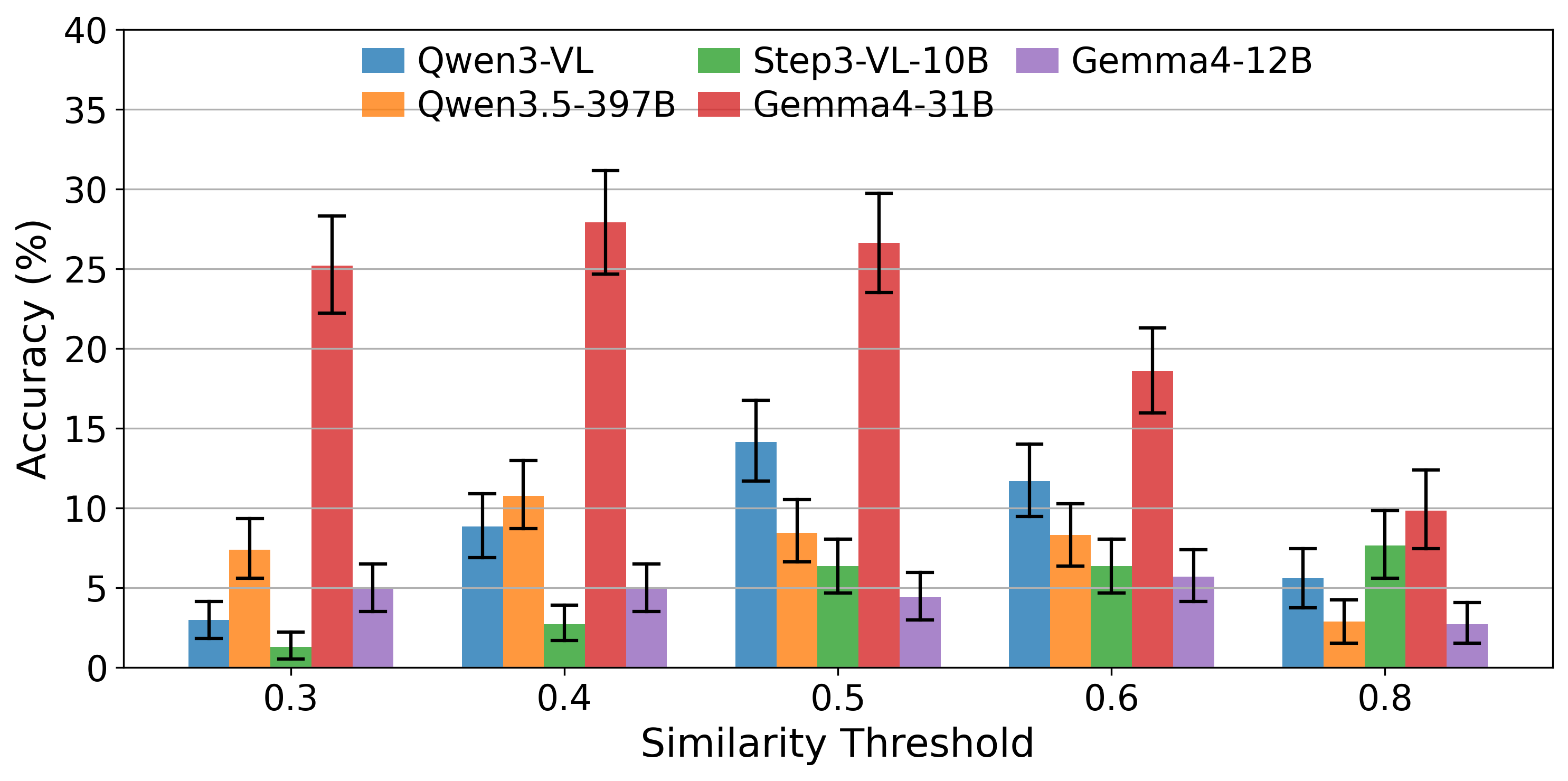}
  \caption{Verified Correct accuracy with 95\% confidence intervals, showing sampling uncertainty around evidence-grounded success.}
  \label{fig:all_models_accuracies_with_ci}
\end{figure*}

Figure~\ref{fig:all_models_accuracies_with_ci} and Table~\ref{tab:verified_accuracy_ci_all_models} show that \textbf{Gemma4-31B}'s verified-accuracy advantage is large relative to sampling uncertainty from $\tau=0.3$ to $\tau=0.6$. For \textbf{Qwen3-VL-235B-A22B-Instruct-FP8}, the intervals show that its verified edge over the other non-Gemma4-31B models is clearest at $\tau=0.5$: at $\tau=0.4$, its interval overlaps with Qwen3.5-397B, while at $\tau=0.5$ it separates from the other non-Gemma4-31B models and at $\tau=0.6$ it remains highest by point estimate. At $\tau=0.8$, the intervals show a different regime: Step3-VL-10B reaches its strongest verified point while Gemma4-31B declines from its low-to-middle-threshold peak, narrowing the verified-accuracy gap in the smallest-gallery setting. Relative to the random-chance baselines in Table~\ref{tab:released_benchmark_examples}, Unverified Overall Accuracy is above chance for most models from $\tau=0.3$ to $\tau=0.6$ and for both \textbf{Gemma4-31B} and \textbf{Step3-VL-10B} at $\tau=0.8$, whereas the lower Verified Correct rates show that above-chance final-answer accuracy does not necessarily imply verification-grounded success.

\begin{table}[htbp]
\centering
\small
\setlength{\tabcolsep}{2pt}
\renewcommand{\arraystretch}{0.95}
\resizebox{\textwidth}{!}{%
\begin{tabular}{@{}l*{15}{r}@{}}
\toprule
 & \multicolumn{5}{c}{\textbf{Verified}} & \multicolumn{5}{c}{\textbf{R.G.}} & \multicolumn{5}{c}{\textbf{Unver. overall}} \\
\cmidrule(lr){2-6}\cmidrule(lr){7-11}\cmidrule(l){12-16}
\textbf{Model} & .3 & .4 & .5 & .6 & .8 & .3 & .4 & .5 & .6 & .8 & .3 & .4 & .5 & .6 & .8 \\
\midrule
Qwen3-VL & 3.0 & 8.8 & 14.2 & 11.7 & 5.6 & 3.8 & 7.0 & 8.4 & 9.4 & 7.1 & 6.8 & 15.8 & 22.6 & 21.0 & 12.7 \\
Qwen3.5-397B & 7.4 & 10.8 & 8.4 & 8.3 & 2.9 & 1.2 & 1.9 & 3.6 & 4.8 & 7.3 & 8.6 & 12.7 & 12.1 & 13.1 & 10.2 \\
Step3-VL-10B & 1.3 & 2.7 & 6.4 & 6.4 & 7.6 & 2.3 & 4.7 & 4.2 & 7.8 & 11.4 & 3.6 & 7.4 & 10.5 & 14.2 & 19.0 \\
Gemma4-31B & 25.2 & 27.9 & 26.6 & 18.6 & 9.8 & 10.9 & 14.0 & 15.1 & 19.7 & 15.1 & 36.1 & 41.9 & 41.7 & 38.3 & 25.0 \\
Gemma4-12B & 4.9 & 4.9 & 4.4 & 5.7 & 2.7 & 1.9 & 2.6 & 3.0 & 3.4 & 5.1 & 6.9 & 7.5 & 7.4 & 9.1 & 7.8 \\
\bottomrule
\end{tabular}%
}
\caption{Threshold-level accuracy metrics for all models (\%). Column labels .3--.8 denote similarity thresholds. R.G. denotes Random-Guess Correct accuracy, and Unver. overall denotes Verified Correct plus Random-Guess Correct accuracy.}
\label{tab:per_threshold_accuracies_all_models}
\end{table}

\begin{table}[htbp]
\centering
\small
\setlength{\tabcolsep}{3pt}
\renewcommand{\arraystretch}{0.95}
\resizebox{\textwidth}{!}{%
\begin{tabular}{@{}lccccc@{}}
\toprule
\textbf{Model} & $\boldsymbol{\tau=.3}$ & $\boldsymbol{\tau=.4}$ & $\boldsymbol{\tau=.5}$ & $\boldsymbol{\tau=.6}$ & $\boldsymbol{\tau=.8}$ \\
\midrule
Qwen3-VL & 3.0 [1.8, 4.2] & 8.8 [6.9, 10.9] & 14.2 [11.7, 16.8] & 11.7 [9.5, 14.0] & 5.6 [3.7, 7.5] \\
Qwen3.5-397B & 7.4 [5.6, 9.4] & 10.8 [8.7, 13.0] & 8.4 [6.6, 10.5] & 8.3 [6.4, 10.3] & 2.9 [1.5, 4.2] \\
Step3-VL-10B & 1.3 [0.5, 2.2] & 2.7 [1.7, 3.9] & 6.4 [4.7, 8.1] & 6.4 [4.7, 8.1] & 7.6 [5.6, 9.8] \\
Gemma4-31B & 25.2 [22.2, 28.3] & 27.9 [24.7, 31.2] & 26.6 [23.5, 29.7] & 18.6 [16.0, 21.3] & 9.8 [7.5, 12.4] \\
Gemma4-12B & 4.9 [3.5, 6.5] & 4.9 [3.5, 6.5] & 4.4 [3.0, 6.0] & 5.7 [4.2, 7.4] & 2.7 [1.5, 4.1] \\
\bottomrule
\end{tabular}%
}
\caption{Verified Correct accuracy with 95\% bootstrap confidence intervals across similarity thresholds for all models (\%). Each cell reports mean [lower, upper].}
\label{tab:verified_accuracy_ci_all_models}
\end{table}

\subsection{Efficiency}
\begin{figure}[!htbp]
  \centering
  \includegraphics[width=0.95\textwidth]{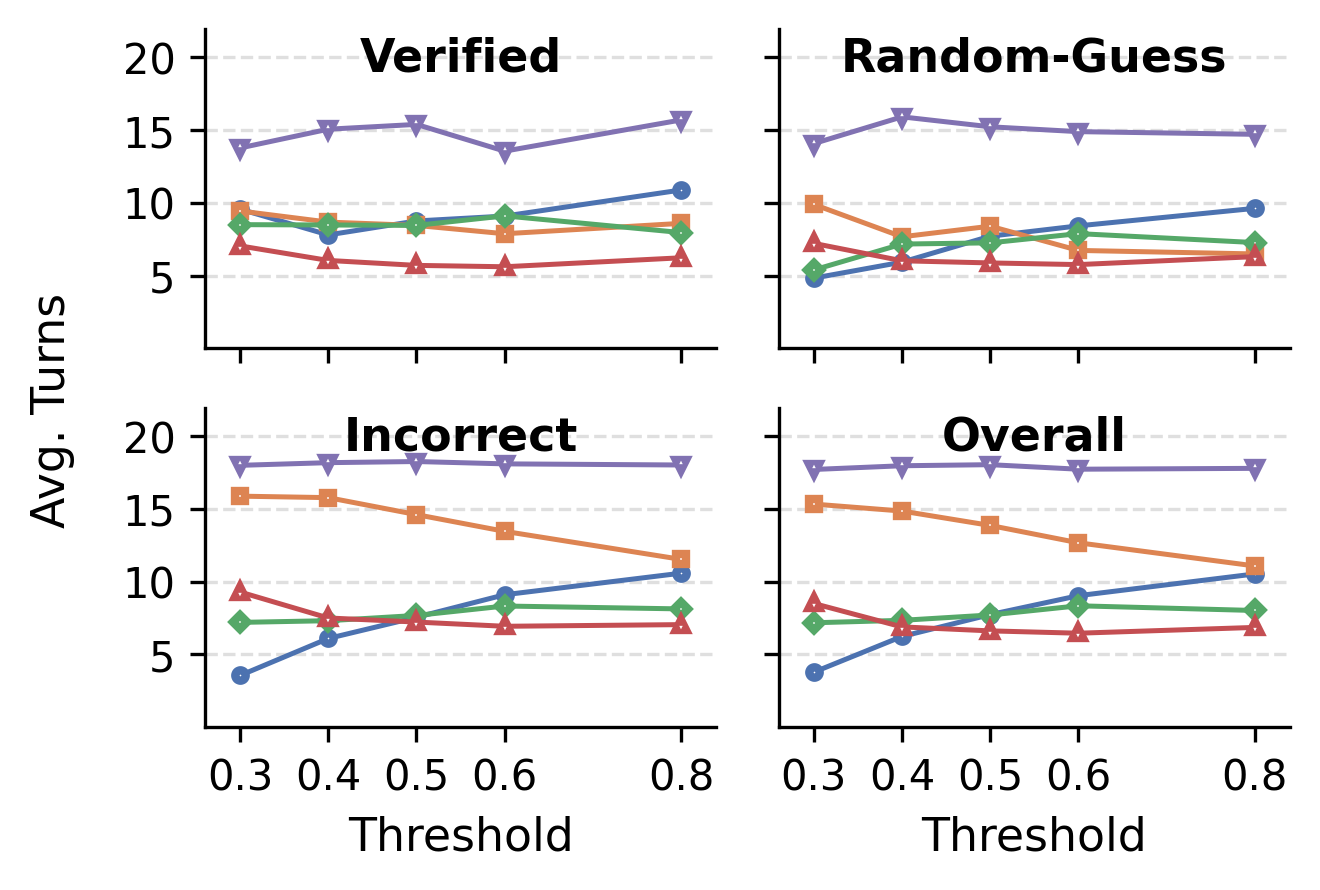}
  \modelcolorlegend
  \caption{Average post-upload turns by outcome category and similarity threshold.}
  \label{fig:all_models_average_rounds_main}
\end{figure}

We measure interaction efficiency by the number of post-upload dialogue turns used before a final decision or exhaustion of the 20-question budget. Figure~\ref{fig:all_models_average_rounds_main} and Table~\ref{tab:per_threshold_rounds_all_models} report average turns by outcome category, since a single overall average can hide different stopping behavior for Verified Correct, Random-Guess Correct, and incorrect trajectories.

The outcome-conditioned patterns reveal several distinct efficiency profiles. \textbf{Gemma4-31B} keeps verified and random-guess-correct traces compact (5.6--7.2 rounds across thresholds), while incorrect traces are only moderately longer (6.9--9.3), indicating efficient stopping when evidence is sufficient. \textbf{Step3-VL-10B} is comparatively stable across outcome categories, with verified, incorrect, and overall averages staying in a narrow 7--9-round range and random-guess-correct traces often slightly shorter. \textbf{Gemma4-12B} uses long traces across all outcomes, with incorrect and overall trajectories near the 18-round budget across all thresholds, showing persistence without a corresponding gain in verified success. \textbf{Qwen3.5-397B-A17B-FP8} shows a sharper outcome split: verified and random-guess-correct traces are relatively compact (6.5--9.9 rounds), but incorrect traces average 11.5--15.9 rounds, so its overall average is largely shaped by failed long dialogues. \textbf{Qwen3-VL-235B-A22B-Instruct-FP8} exhibits pronounced threshold dependence: at $\tau=0.3$, incorrect trajectories terminate prematurely (3.6 rounds on average), compressing the overall average to 3.8, whereas verified traces remain substantially longer at 9.6; as thresholds increase, incorrect and overall trajectories converge toward the verified-trace length, reaching 10.5--10.9 rounds at $\tau=0.8$, reflecting a shift from premature commitment in broad-gallery settings to more sustained interaction as galleries contract.

These patterns show that lower turn counts are not always better. Compact verified traces indicate efficient evidence use, but short incorrect or random-guess-correct traces can indicate early commitment without verification-passing support. Conversely, long incorrect traces indicate continued interaction without effective candidate reduction.

Table~\ref{tab:budget_exhaustion} reports the budget-exhaustion rate: the fraction of trajectories that use all 20 questions without producing a final guess. \textbf{Gemma4-12B} and \textbf{Qwen3.5-397B-A17B-FP8} stand out, with exhaustion rates exceeding 50\% at several thresholds. For Qwen3.5-397B, the rate rises sharply with $\tau$, from 28.9\% at $\tau=0.3$ to 54.8\% at $\tau=0.8$, consistent with its long incorrect traces and frequent invalid-question loops (analyzed in Section~\ref{sec:protocol_compliance}). Gemma4-12B is consistently above 52\% across all thresholds, reflecting persistent interaction without convergence. In contrast, \textbf{Qwen3-VL-235B-A22B-Instruct-FP8} almost never exhausts the budget (at most 0.3\%), consistent with its early-stopping profile, while \textbf{Gemma4-31B} and \textbf{Step3-VL-10B} remain below 8\%, indicating that they generally commit to a final guess within the allotted budget even when incorrect.

\begin{table}[htbp]
\centering
\small
\setlength{\tabcolsep}{0.8pt}
\renewcommand{\arraystretch}{0.95}
\resizebox{\textwidth}{!}{%
\begin{tabular}{@{}l*{20}{r}@{}}
\toprule
 & \multicolumn{5}{c}{\textbf{Verified}} & \multicolumn{5}{c}{\textbf{R.G.}} & \multicolumn{5}{c}{\textbf{Incorrect}} & \multicolumn{5}{c}{\textbf{Overall}} \\
\cmidrule(lr){2-6}\cmidrule(lr){7-11}\cmidrule(lr){12-16}\cmidrule(l){17-21}
\textbf{Model} & .3 & .4 & .5 & .6 & .8 & .3 & .4 & .5 & .6 & .8 & .3 & .4 & .5 & .6 & .8 & .3 & .4 & .5 & .6 & .8 \\
\midrule
Qwen3-VL &9.57 & 7.79 & 8.75 & 9.09 & 10.88 & 4.83 & 5.93 & 7.71 & 8.42 & 9.62 & 3.55 & 6.09 & 7.54 & 9.09 & 10.57 & 3.78 & 6.23 & 7.73 & 9.03 & 10.52 \\
Qwen3.5-397B & 9.44 & 8.67 & 8.45 & 7.88 & 8.59 & 9.89 & 7.67 & 8.39 & 6.73 & 6.49 & 15.86 & 15.76 & 14.59 & 13.44 & 11.52 & 15.32 & 14.84 & 13.85 & 12.66 & 11.07 \\
Step3-VL-10B & 8.50 & 8.48 & 8.45 & 9.10 & 7.96 & 5.39 & 7.17 & 7.25 & 7.88 & 7.27 & 7.19 & 7.31 & 7.67 & 8.32 & 8.12 & 7.17 & 7.34 & 7.70 & 8.33 & 8.01 \\
Gemma4-31B & 7.04 & 6.04 & 5.70 & 5.61 & 6.22 & 7.21 & 6.02 & 5.87 & 5.76 & 6.30 & 9.28 & 7.50 & 7.22 & 6.93 & 7.04 & 8.49 & 6.88 & 6.61 & 6.45 & 6.85 \\
Gemma4-12B & 13.76 & 15.05 & 15.38 & 13.55 & 15.69 & 14.07 & 15.90 & 15.22 & 14.88 & 14.70 & 17.98 & 18.16 & 18.24 & 18.08 & 18.00 & 17.69 & 17.95 & 18.02 & 17.71 & 17.77 \\
\bottomrule
\end{tabular}%
}
\caption{Average post-upload turns per similarity threshold for all models, stratified by outcome category. Column labels .3--.8 denote similarity thresholds, and R.G. denotes Random-Guess Correct.}
\label{tab:per_threshold_rounds_all_models}
\end{table}

\begin{table}[htbp]
\centering
\small
\setlength{\tabcolsep}{4pt}
\renewcommand{\arraystretch}{1.0}
\begin{tabular}{@{}lrrrrr@{}}
\toprule
\textbf{Model} & $\boldsymbol{\tau=0.3}$ & $\boldsymbol{\tau=0.4}$ & $\boldsymbol{\tau=0.5}$ & $\boldsymbol{\tau=0.6}$ & $\boldsymbol{\tau=0.8}$ \\
\midrule
Qwen3-VL & 0.2 & 0.0 & 0.3 & 0.0 & 0.0 \\
Qwen3.5-397B & 28.9 & 33.9 & 44.0 & 53.8 & 54.8 \\
Step3-VL-10B & 2.2 & 4.9 & 5.5 & 4.5 & 7.9 \\
Gemma4-31B & 7.0 & 3.6 & 2.7 & 3.5 & 6.5 \\
Gemma4-12B & 53.7 & 52.6 & 56.5 & 56.4 & 55.3 \\
\bottomrule
\end{tabular}
\caption{Budget-exhaustion rate (\%): fraction of trajectories that use all 20 questions without producing a final guess, by model and similarity threshold.}
\label{tab:budget_exhaustion}
\end{table}
\subsection{Protocol compliance}
\label{sec:protocol_compliance}
\begin{figure}[!htbp]
  \centering
  \includegraphics[width=0.95\textwidth]{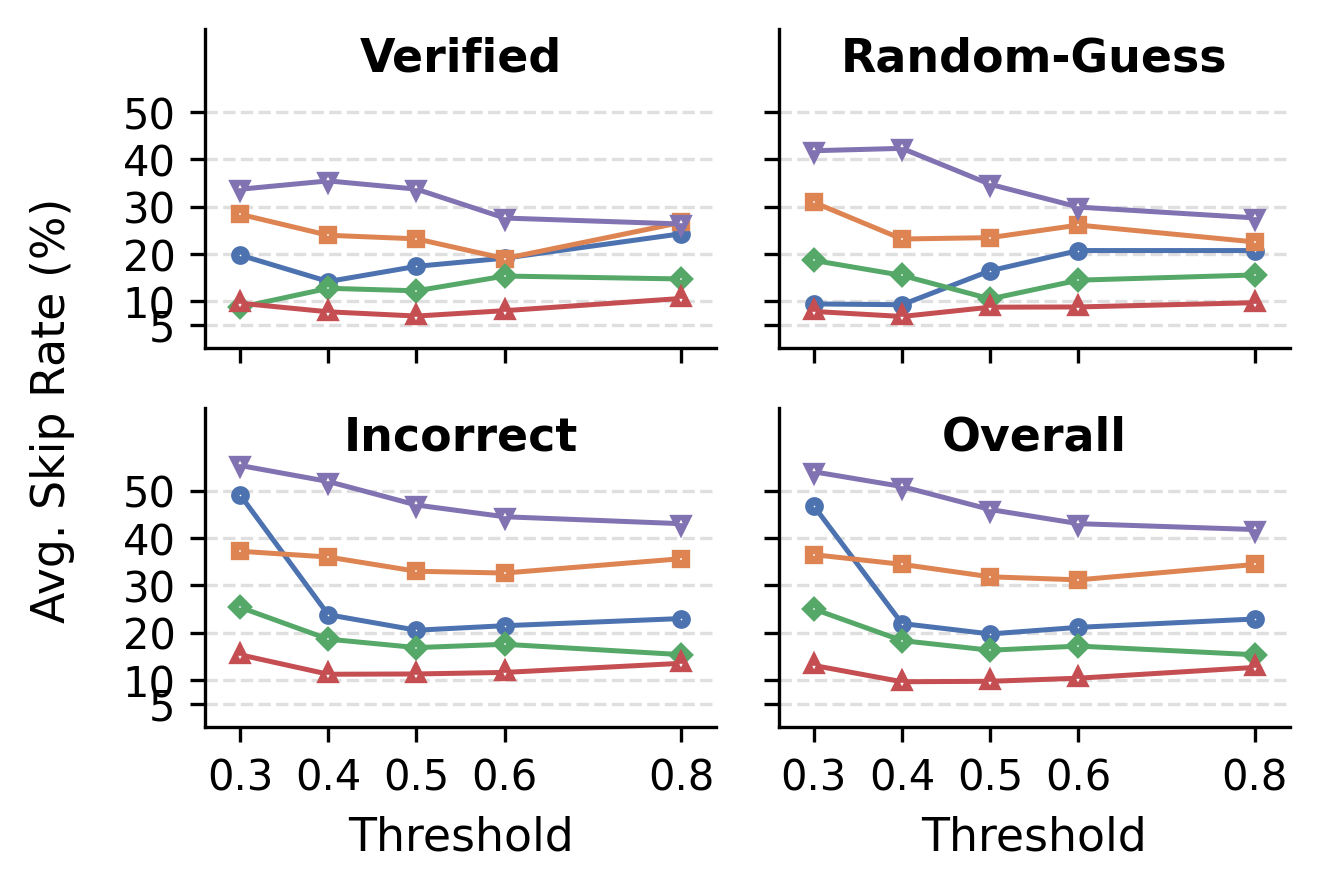}
  \modelcolorlegend
  \caption{Post-upload \texttt{Skip} rates by outcome category and similarity threshold.}
  \label{fig:all_models_average_skip_rates_main}
\end{figure}

\begin{figure}[p]
  \centering
  \begin{subfigure}[t]{0.95\textwidth}
    \centering
    \includegraphics[width=\linewidth]{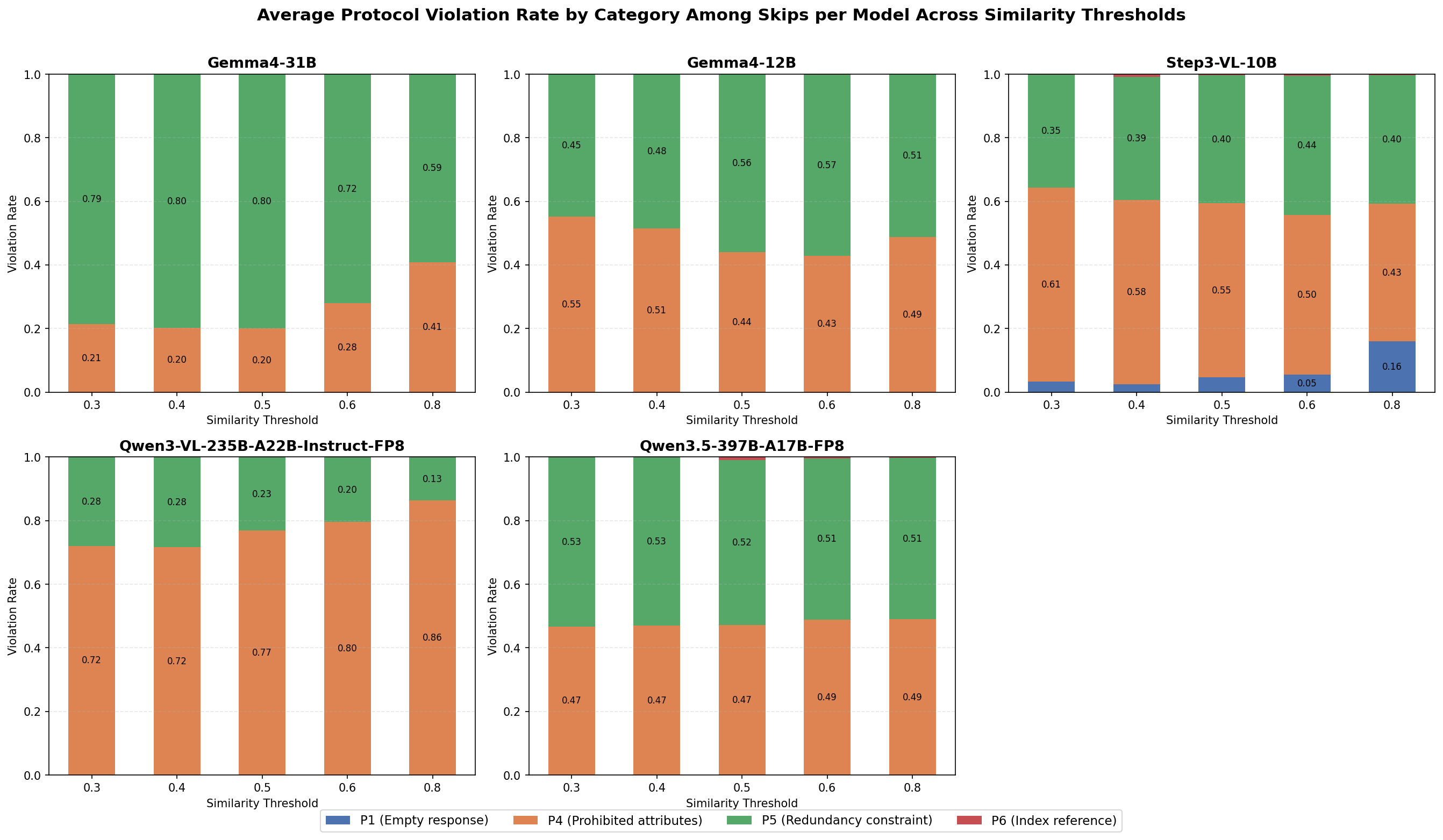}
    \caption{Share of skipped turns assigned to each rubric category (P1, P4, P5, P6) by model and threshold.}
    \label{fig:average_protocol_violation_rate_by_category}
  \end{subfigure}
  \vspace{0.5em}
  \begin{subfigure}[t]{0.95\textwidth}
    \centering
    \includegraphics[width=\linewidth]{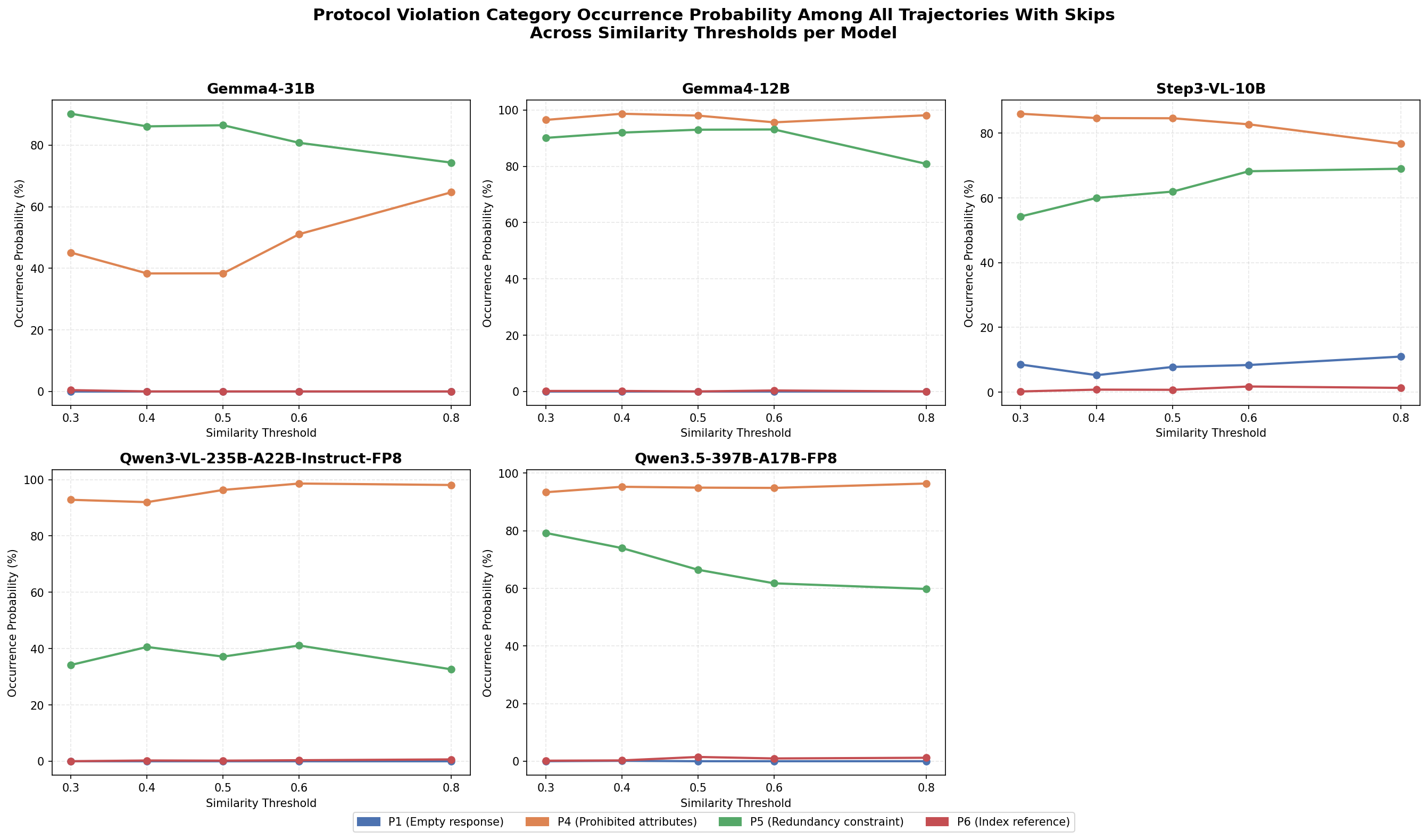}
    \caption{Fraction of trajectories with at least one skip in each rubric category by model and threshold.}
    \label{fig:protocol_violation_occurrence_from_total}
  \end{subfigure}
  \caption{Protocol-rubric breakdown for post-upload \texttt{Skip} responses. (a) shows the per-turn composition across rubric categories, while (b) reports trajectory-level occurrence.}
  \label{fig:protocol_violation_breakdown}
\end{figure}

\begin{figure}[!t]
  \centering
  \includegraphics[width=\linewidth]{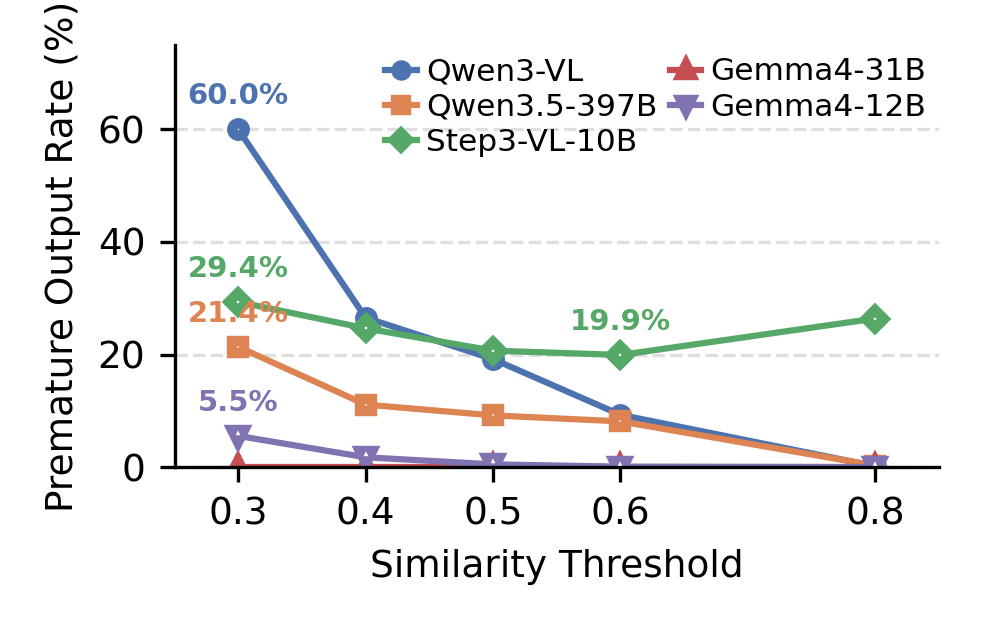}
  \caption{Premature-output rates before the required \texttt{End of uploading} signal.}
  \label{fig:premature_output_main}
\end{figure}

Protocol compliance measures whether a model follows the AMIGO interaction contract independently of final-answer correctness. We analyze post-upload \texttt{Skip} responses in Figure~\ref{fig:all_models_average_skip_rates_main} and Table~\ref{tab:per_threshold_skip_rate_all_models}, premature outputs in Figure~\ref{fig:premature_output_main} and Table~\ref{tab:pre_violation_all_models}, and post-upload violation categories in Figure~\ref{fig:protocol_violation_breakdown}.

The compliance profiles separate post-upload invalid questions from premature upload-phase responses. \textbf{Gemma4-31B} is consistently clean on both dimensions, combining low post-upload \texttt{Skip} rates with no premature outputs. \textbf{Step3-VL-10B} asks fewer invalid post-upload questions than most models but persistently violates the upload-completion boundary. \textbf{Qwen3-VL-235B-A22B-Instruct-FP8} diverges across categories: its post-upload \texttt{Skip} behavior is much worse on incorrect traces than on successful traces in the broadest-gallery setting, while its premature-output rate drops sharply from $\tau=0.3$ to $\tau=0.4$ and continues decreasing afterward. \textbf{Qwen3.5-397B-A17B-FP8} and \textbf{Gemma4-12B} mainly struggle with high post-upload invalid-question rates, with Gemma4-12B otherwise showing strong upload-phase discipline.

Figure~\ref{fig:all_models_average_skip_rates_main} and Table~\ref{tab:per_threshold_skip_rate_all_models} show substantial variation in post-upload \texttt{Skip} behavior across outcome categories. \textbf{Gemma4-31B} maintains the lowest overall skip rate at every threshold (9.6--13.0\%) and remains consistently low across verified, random-guess-correct, and incorrect trajectories, indicating the most stable post-upload protocol adherence. \textbf{Step3-VL-10B} also keeps post-upload skip rates relatively controlled after $\tau=0.3$, with overall rates between 15.3\% and 18.3\% from $\tau=0.4$ to $\tau=0.8$. \textbf{Qwen3-VL-235B-A22B-Instruct-FP8} is category-divergent: at $\tau=0.3$, its overall skip rate reaches 46.8\% due to a 49.1\% rate on incorrect traces, while verified and random-guess-correct traces remain moderate (19.7\% and 9.4\%); from $\tau=0.4$ onward, the overall rate stabilizes at 19.7--22.9\%, though incorrect traces consistently skip more than successful ones. \textbf{Qwen3.5-397B-A17B-FP8} and \textbf{Gemma4-12B} have higher skip rates across outcome categories (overall 31.1--36.5\% and 41.8--54.0\%, respectively), indicating frequent invalid or redundant questions after the gallery is available.

\begin{table}[htbp]
\centering
\small
\setlength{\tabcolsep}{1.5pt}
\renewcommand{\arraystretch}{0.95}
\resizebox{\textwidth}{!}{%
\begin{tabular}{@{}l*{20}{r}@{}}
\toprule
 & \multicolumn{5}{c}{\textbf{Verified}} & \multicolumn{5}{c}{\textbf{R.G.}} & \multicolumn{5}{c}{\textbf{Incorrect}} & \multicolumn{5}{c}{\textbf{Overall}} \\
\cmidrule(lr){2-6}\cmidrule(lr){7-11}\cmidrule(lr){12-16}\cmidrule(l){17-21}
\textbf{Model} & .3 & .4 & .5 & .6 & .8 & .3 & .4 & .5 & .6 & .8 & .3 & .4 & .5 & .6 & .8 & .3 & .4 & .5 & .6 & .8 \\
\midrule
Qwen3-VL &19.7 & 14.1 & 17.4 & 19.1 & 24.2 & 9.4 & 9.2 & 16.4 & 20.7 & 20.7 & 49.1 & 23.8 & 20.5 & 21.4 & 23.0 & 46.8 & 21.9 & 19.7 & 21.1 & 22.9 \\
Qwen3.5-397B & 28.4 & 23.9 & 23.2 & 19.0 & 26.6 & 30.9 & 23.1 & 23.4 & 26.1 & 22.5 & 37.2 & 36.0 & 33.0 & 32.6 & 35.6 & 36.5 & 34.4 & 31.8 & 31.1 & 34.4 \\
Step3-VL-10B & 8.7 & 12.7 & 12.2 & 15.3 & 14.7 & 18.6 & 15.4 & 10.5 & 14.4 & 15.5 & 25.3 & 18.6 & 16.8 & 17.5 & 15.3 & 25.0 & 18.3 & 16.3 & 17.1 & 15.3 \\
Gemma4-31B & 9.6 & 7.7 & 6.8 & 8.0 & 10.6 & 7.8 & 6.8 & 8.7 & 8.8 & 9.7 & 15.3 & 11.2 & 11.2 & 11.6 & 13.5 & 13.0 & 9.6 & 9.7 & 10.4 & 12.6 \\
Gemma4-12B & 33.7 & 35.4 & 33.7 & 27.6 & 26.3 & 41.8 & 42.3 & 34.7 & 29.9 & 27.6 & 55.3 & 51.9 & 47.0 & 44.5 & 43.0 & 54.0 & 50.9 & 46.0 & 43.0 & 41.8 \\
\bottomrule
\end{tabular}%
}
\caption{Average post-upload \texttt{Skip} rate per similarity threshold for all models (\%), stratified by outcome category. Column labels .3--.8 denote similarity thresholds, and R.G. denotes Random-Guess Correct.}
\label{tab:per_threshold_skip_rate_all_models}
\end{table}

\paragraph{Post-upload violation-category breakdown}
The post-upload violation-category breakdown in Figure~\ref{fig:protocol_violation_breakdown} uses the protocol-compliance rubric to localize the invalid-action failures behind post-upload \texttt{Skip} responses. Under this rubric, a \texttt{Skip} is assigned to a concrete rule dimension rather than treated as a generic failed turn: P1 covers malformed or non-answerable questions, P4 covers prohibited-attribute questions, P5 covers redundant attribute re-enumeration, and P6 covers image-index references. Across these diagnostics, P4 and P5 dominate (Figure~\ref{fig:protocol_violation_occurrence_from_total}), indicating that most invalid turns arise from substantive violations of the attribute-query rubric rather than simple formatting errors.

The per-model composition (Figure~\ref{fig:average_protocol_violation_rate_by_category}) reveals distinct violation signatures. \textbf{Qwen3-VL-235B-A22B-Instruct-FP8} is strongly P4-dominant (72--86\% of skipped turns), with P5 redundancy declining from 28\% to 13\%. \textbf{Qwen3.5-397B-A17B-FP8} remains roughly balanced between P4 and P5, indicating both prohibited-attribute inquiries and redundant re-enumeration as recurring failure modes. \textbf{Gemma4-12B} is similarly mixed, though its higher overall skip rate means these violations occur more frequently per trajectory. \textbf{Gemma4-31B} skews P5-heavy at low-to-middle thresholds, suggesting that its rare violations stem primarily from redundant re-enumeration rather than prohibited attributes, before shifting toward a more balanced profile at $\tau=0.8$. \textbf{Step3-VL-10B} maintains a relatively even P1/P4/P5 composition, with its P1 share increasing at the strictest threshold, reflecting occasional formatting or answerability failures alongside substantive rubric violations.

\FloatBarrier

Premature outputs capture a different protocol failure: asking or guessing before the required \texttt{End of uploading} signal. Figure~\ref{fig:premature_output_main} and Table~\ref{tab:pre_violation_all_models} show that \textbf{Gemma4-31B} has a 0.0\% premature-output rate and \textbf{Gemma4-12B} is nearly clean after $\tau=0.3$. \textbf{Qwen3-VL-235B-A22B-Instruct-FP8} is highly threshold-sensitive on this metric, falling from 60.0\% at $\tau=0.3$ to 26.5\% at $\tau=0.4$ and then to 0.2\% at $\tau=0.8$. \textbf{Step3-VL-10B} remains persistently affected, staying between 19.9\% and 29.4\% across thresholds, making upload-phase turn-taking its clearest protocol-compliance weakness. \textbf{Qwen3.5-397B-A17B-FP8} improves steadily from 21.4\% at $\tau=0.3$ to 0.3\% at $\tau=0.8$.

Table~\ref{tab:pre_violation_all_models} reports the exact premature-output rates by model and threshold.

\begin{table}[htbp]
\centering
\small
\setlength{\tabcolsep}{4pt}
\renewcommand{\arraystretch}{1.0}
\begin{tabular}{@{}lrrrrr@{}}
\toprule
\textbf{Model} & $\boldsymbol{\tau=0.3}$ & $\boldsymbol{\tau=0.4}$ & $\boldsymbol{\tau=0.5}$ & $\boldsymbol{\tau=0.6}$ & $\boldsymbol{\tau=0.8}$ \\
\midrule
Qwen3-VL & 60.0 & 26.5 & 19.2 & 9.3 & 0.2 \\
Qwen3.5-397B & 21.4 & 11.1 & 9.2 & 8.1 & 0.3 \\
Step3-VL-10B & 29.4 & 24.6 & 20.7 & 19.9 & 26.4 \\
Gemma4-31B & 0.0 & 0.0 & 0.0 & 0.0 & 0.0 \\
Gemma4-12B & 5.5 & 1.8 & 0.5 & 0.1 & 0.0 \\
\bottomrule
\end{tabular}
\caption{Premature-output rate before the required \texttt{End of uploading} signal per similarity threshold for all models (\%).}
\label{tab:pre_violation_all_models}
\end{table}

\subsection{Robustness under controlled noisy feedback}
Robustness under controlled noisy feedback evaluates whether an interactive model can recover when one piece of oracle feedback is deliberately corrupted while the rest of the evidence trail remains unchanged. This matters because AMIGO questions often target fine-grained or subjective visual attributes, where an oracle response may occasionally be locally ambiguous, inconsistent with later evidence, or simply wrong. A reliable agent should not treat every individual response as irrevocable; it should maintain competing hypotheses, use later questions to repair the candidate set, and still converge when the dialogue contains a small amount of controlled noise.

We instantiate the noisy feedback with a controlled perturbation: one random \emph{Yes}/\emph{No} oracle response is flipped in each eligible trajectory, the earlier dialogue prefix is preserved, and the model continues from the contradictory feedback. We repeat five trials per setting. \emph{Unsure} and \texttt{Skip} responses are not flipped, isolating injected binary contradiction rather than protocol noise.

Table~\ref{tab:recovery_rate_noisy_feedback_main} reports threshold-level recovery rates and the macro-average recovery rate under this noisy-feedback setting. The macro-average is the arithmetic mean of the per-threshold recovery rates. \textbf{Gemma4-31B} is strongest by macro average, with a 59.1\% macro-average recovery rate. \textbf{Qwen3-VL-235B-A22B-Instruct-FP8} has the second-highest macro-average recovery rate at 51.5\%, followed by \textbf{Qwen3.5-397B-A17B-FP8} at 37.4\%, \textbf{Gemma4-12B} at 33.8\%, and \textbf{Step3-VL-10B} at 29.9\%.

The per-threshold recovery rate can vary substantially for each model. \textbf{Gemma4-31B} remains consistently high across thresholds, staying within 54.5--64.6\%, which indicates stable repair rather than a single favorable setting. \textbf{Qwen3-VL-235B-A22B-Instruct-FP8} improves from 49.6\% at $\tau=0.3$ to 58.7\% at $\tau=0.6$ but drops to 43.0\% at $\tau=0.8$, where it falls slightly below Step3-VL-10B. \textbf{Qwen3.5-397B-A17B-FP8} and \textbf{Gemma4-12B} generally decline as $\tau$ increases. \textbf{Step3-VL-10B} shows a different pattern: weak recovery at low-$\tau$, larger-gallery settings (18.0\% at $\tau=0.3$ and 26.7\% at $\tau=0.4$) pulls down its macro average, even though its best recovery occurs in the highest-$\tau$, smaller-gallery setting (45.3\% at $\tau=0.8$).

\begin{table}[!t]
  \centering
  \small
  \setlength{\tabcolsep}{2.5pt}
  \renewcommand{\arraystretch}{0.95}
  \begin{tabular}{@{}lcccccc c@{}}
    \toprule
    \textbf{Model} & $\boldsymbol{\tau{=}0.3}$ & $\boldsymbol{\tau{=}0.4}$ & $\boldsymbol{\tau{=}0.5}$ & $\boldsymbol{\tau{=}0.6}$ & $\boldsymbol{\tau{=}0.8}$ & \textbf{Macro avg.} \\
    \midrule
    Gemma4-31B & $64.6{\pm}3.0$ & $61.4{\pm}2.6$ & $55.2{\pm}4.4$ & $59.9{\pm}3.2$ & $54.5{\pm}0.9$ & $59.1{\pm}2.8$ \\
    Qwen3-VL & $49.6{\pm}10.9$ & $58.2{\pm}5.5$ & $48.1{\pm}3.2$ & $58.7{\pm}5.1$ & $43.0{\pm}9.7$ & $51.5{\pm}6.9$ \\
    Qwen3.5-397B & $43.5{\pm}11.1$ & $39.8{\pm}5.3$ & $34.6{\pm}0.9$ & $38.8{\pm}9.1$ & $30.6{\pm}10.5$ & $37.4{\pm}7.4$ \\
    Step3-VL-10B & $18.0{\pm}14.8$ & $26.7{\pm}8.7$ & $32.7{\pm}4.8$ & $26.9{\pm}4.9$ & $45.3{\pm}7.5$ & $29.9{\pm}8.1$ \\
    Gemma4-12B & $45.3{\pm}6.8$ & $31.6{\pm}8.3$ & $33.5{\pm}7.9$ & $31.4{\pm}3.0$ & $27.5{\pm}14.4$ & $33.8{\pm}8.1$ \\
    \bottomrule
  \end{tabular}
  \caption{Threshold-level and macro-average recovery rates (\%) under controlled noisy feedback induced by one deliberately flipped Yes/No oracle response. Entries are mean $\pm$ standard deviation over five runs; Macro avg. averages the per-threshold means and standard deviations across thresholds.}
  \label{tab:recovery_rate_noisy_feedback_main}
\end{table}

Together, these results show that recovery from a single corrupted response exposes distinct profiles: Gemma4-31B pairs the highest macro average with cross-threshold stability, Qwen3-VL's second-ranked macro conceals a sharp rise-then-drop, and Step3-VL-10B's low macro hides strong recovery at the toughest threshold. Notably, these recovery profiles track the same threshold-dependent ranking as Verified Correct accuracy---Gemma4-31B leads at low-to-middle $\tau$, Qwen3-VL peaks at middle $\tau$, and Step3-VL-10B is strongest at $\tau=0.8$---suggesting that the ability to ground evidence and the ability to repair corrupted feedback share an underlying dependence on effective candidate reduction.

\subsection{Additional Reasoning-Mode Comparison}

Table~\ref{tab:tau08_additional_reasoning_results} compares two additional $\tau=0.8$ inference settings with two main-run references: \textbf{Gemma4-31B} with thinking enabled, \textbf{Muse-Glimmer-30B} with default low-effort reasoning, \textbf{Step3-VL-10B} with default thinking, and \textbf{Gemma4-31B} without thinking. Because these settings are evaluated only at $\tau=0.8$, they should be interpreted as targeted comparisons rather than cross-threshold model rankings.

\begin{table}[t]
\centering
\small
\setlength{\tabcolsep}{2pt}
\resizebox{\textwidth}{!}{%
\begin{tabular}{@{}lccccccc@{}}
\toprule
\textbf{Model/setting}
& \multicolumn{3}{c}{\textbf{ID accuracy (\%)}}
& \textbf{Efficiency}
& \multicolumn{2}{c}{\textbf{Protocol}}
& \textbf{Robustness} \\
\cmidrule(lr){2-4}\cmidrule(lr){5-5}\cmidrule(lr){6-7}\cmidrule(l){8-8}
& \textbf{Verified} $\uparrow$
& \textbf{R.G.}
& \textbf{Unver. overall} $\uparrow$
& \textbf{Rounds} $\downarrow$
& \textbf{Skip (\%)} $\downarrow$
& \textbf{Prem. (\%)} $\downarrow$
& \textbf{Recov. (\%)} $\uparrow$ \\
\midrule
Gemma4-31B + thinking & 7.5 & 14.9 & 22.4 & 7.8 & 13.0 & 3.9 & 50.0 $\pm$ 0.8 \\
Muse-Glimmer-30B (low) & 11.5 & 16.5 & 28.0 & 10.9 & 28.5 & 77.6 & 51.2 $\pm$ 0.7 \\
Step3-VL-10B, default thinking & 7.6 & 11.4 & 19.0 & 8.01 & 15.3 & 26.4 & 45.3 $\pm$ 7.5 \\
Gemma4-31B, no thinking & 9.8 & 15.1 & 25.0 & 6.85 & 12.63 & 0.0 & 54.5 $\pm$ 0.9 \\
\bottomrule
\end{tabular}%
}
\caption{Additional $\tau=0.8$ results across AMIGO metric axes. Values are percentages except average rounds. R.G. denotes Random-Guess Correct accuracy, Unver. overall denotes Unverified Overall Accuracy, Prem. denotes premature-output rate, and Recov. denotes recovery rate reported as mean $\pm$ standard deviation over five perturbation runs. Arrows indicate the desired direction: $\uparrow$ higher is better, $\downarrow$ lower is better.}
\label{tab:tau08_additional_reasoning_results}
\end{table}

\textbf{Muse-Glimmer-30B} has the highest $\tau=0.8$ identification rates in this comparison, with 11.5\% Verified Correct accuracy and 28.0\% Unverified Overall Accuracy. Its interaction reliability is much weaker: the \texttt{Skip} rate is 28.5\%, and the premature-output rate is 77.6\%. Thus, stronger final identification at this threshold does not imply stronger protocol compliance.

For \textbf{Gemma4-31B}, enabling thinking mode does not improve the $\tau=0.8$ result. Compared with no thinking, thinking lowers Verified Correct accuracy from 9.8\% to 7.5\%, lowers Unverified Overall Accuracy from 25.0\% to 22.4\%, increases average rounds from 6.85 to 7.8, introduces premature outputs, and lowers recovery from 54.5\% to 50.0\%. In this setting, the extra reasoning budget does not translate into better evidence-grounded identification or robustness.

\textbf{Step3-VL-10B}'s default-thinking result serves as a useful reference point. Its 7.6\% Verified Correct accuracy is close to Gemma4-31B thinking mode, but its 19.0\% Unverified Overall Accuracy and 45.3\% recovery rate are lower than both Gemma4-31B settings. Its protocol behavior is mixed: it is much more compliant than Muse-Glimmer-30B, but still has a 26.4\% premature-output rate.

Because Muse-Glimmer-30B and Gemma4-31B with thinking mode are evaluated only at $\tau=0.8$, their recovery values should be interpreted as targeted stress-test results rather than cross-threshold robustness averages. Muse-Glimmer-30B recovers at 51.2\%, above Gemma4-31B with thinking mode and Step3-VL-10B but below Gemma4-31B without thinking, suggesting that default or enabled reasoning alone does not improve robustness under contradictory feedback.

\subsection{Cross-metric model summary}
\begin{table}[!t]
  \centering
  \small
  \setlength{\tabcolsep}{3pt}
  \renewcommand{\arraystretch}{1.4}
  \begin{tabularx}{\textwidth}{@{}l X X X X@{}}
    \toprule
    \textbf{Model} & \textbf{Accuracy} & \textbf{Efficiency} & \textbf{Compliance} & \textbf{Robustness} \\
    \midrule
    Gemma4-31B & \textbf{Leading}~Peaks at mid-$\tau$ & \textbf{Leading}~Compact traces; low exhaustion & \textbf{Leading}~Lowest skip; no premature & Highest recovery; stable across $\tau$ \\
    Qwen3-VL & \textbf{Best non-Gemma4-31B at mid-$\tau$}~Peak at $\tau{=}0.5$ & Early termination at $\tau{=}0.3$; lengthens with $\tau$; minimal exhaustion & High skip at $\tau{=}0.3$; premature drops & 2nd-highest macro recovery; drops at $\tau{=}0.8$ \\
    Qwen3.5 & Stable unverified; verified drops at $\tau{=}0.8$ & Long incorrect; high exhaustion & High skip throughout; moderate premature & Recovery declines with $\tau$ \\
    Step3-VL & Improves with $\tau$; best non-Gemma4-31B at $\tau{=}0.8$ & \textbf{Best non-Gemma4-31B}~Stable; low exhaustion & Lowest non-Gemma4-31B skip; persistently premature & Lowest macro recovery; best at $\tau{=}0.8$ \\
    Gemma4-12B & Low; flat & Near-budget traces; highest exhaustion & Highest skip; clean premature after $\tau{=}0.3$ & Recovery declines with $\tau$ \\
    \bottomrule
  \end{tabularx}
  \caption{Cross-metric model summary. Each cell characterizes the model's qualitative profile on that metric axis across similarity thresholds $\tau\in\{0.3,0.4,0.5,0.6,0.8\}$. Quantitative values are reported in the preceding tables and figures.}
  \label{tab:cross_metric_summary}
\end{table}
Table~\ref{tab:cross_metric_summary} summarizes each model's profile across the four metric axes. Across the main five-model evaluation, \textbf{Gemma4-31B} is the most balanced model, combining the strongest verified identification, low post-upload \texttt{Skip} rates, zero premature outputs, compact successful traces, and the highest noisy-feedback recovery. \textbf{Qwen3-VL-235B-A22B-Instruct-FP8} is strongest among the non-Gemma4-31B models on Unverified Overall Accuracy from $\tau=0.4$ to $\tau=0.6$ and on Verified Correct accuracy at $\tau=0.5$ and $\tau=0.6$, and it has the second-highest macro-average recovery. However, its protocol compliance is threshold-sensitive: it has high premature-output rates at low $\tau$ and elevated \texttt{Skip} rates on incorrect traces, particularly at $\tau=0.3$.

Despite having more parameters (\textbf{Qwen3.5-397B-A17B-FP8} vs.\ \textbf{Qwen3-VL-235B-A22B-Instruct-FP8}), Qwen3.5 does not consistently outperform Qwen3-VL across metric axes: although Qwen3.5 has higher Verified Correct accuracy at low $\tau$ and lower premature-output rates, Qwen3-VL achieves higher Verified Correct accuracy at middle thresholds, higher macro-average recovery, lower overall \texttt{Skip} rates at middle and high thresholds, and shorter incorrect traces. Larger scale does not necessarily translate into stronger interactive grounding in this comparison.

\textbf{Step3-VL-10B} improves as galleries become smaller and is strongest among the non-Gemma4-31B models at $\tau=0.8$, but persistent premature outputs weaken its protocol profile. Its low macro-average recovery reflects poor low-threshold recovery rather than uniformly poor recovery. \textbf{Qwen3.5-397B-A17B-FP8} and \textbf{Gemma4-12B} expose a different weakness: both interact for many turns but often accumulate invalid or redundant questions without corresponding verified success. The additional $\tau=0.8$ reasoning-mode comparison further shows that more explicit reasoning does not automatically improve AMIGO performance across evidence grounding, efficiency, compliance, and robustness. Overall, AMIGO separates models that merely reach occasional correct final answers from models that combine verified evidence, efficient dialogue, protocol adherence, and recovery from controlled feedback noise.

\section{Trajectory Structure for Reinforcement Learning}

AMIGO also collects interaction trajectories during evaluation for future multimodal reinforcement learning and preference optimization studies. Each interaction naturally defines a sequential decision process: the state includes the image gallery, dialogue history, and accumulated constraints; the action is a protocol-constrained attribute question or a terminal guess; and the observation is the oracle response, including \texttt{Yes}, \texttt{No}, \texttt{Unsure}, or \texttt{Skip}. Because the evaluation records turn-level outcomes and the verification module tracks candidate-set dynamics, AMIGO captures both terminal outcomes and intermediate reasoning behavior.

This structure could enable dense, automatically computable reward signals. Invalid questions can receive negative rewards through \texttt{Skip}; valid questions can be rewarded according to candidate-set reduction, estimated information gain, or progress toward a singleton feasible set; and terminal rewards can be assigned when the model correctly identifies the target with verification-passing evidence. These signals directly reflect the benchmark's core requirements: asking valid discriminative questions, preserving constraints across turns, and stopping only when the dialogue supports a unique target.

The same structure could support several learning studies. Once rewards are defined and the collected data provide sufficient behavioral coverage, completed traces could support offline reinforcement-learning datasets. Preference optimization can compare alternative questions for the same candidate state and favor questions that are valid, non-redundant, and more informative. These uses build on the state, action, observation, and verification structure.

\section{Conclusion}

AMIGO reframes multi-image evaluation as an interactive grounding problem rather than a final-answer recognition problem. By requiring models to identify a hidden target through constrained Yes/No questions, AMIGO makes question selection, constraint tracking, evidence accumulation, stopping behavior, protocol compliance, and feedback recovery directly measurable. The verification module is central to this framing: it separates correct final guesses from trajectories whose accumulated evidence uniquely supports the target, exposing failures that would be hidden by Unverified Overall Accuracy alone.

The experiments show that current open-source VLMs remain far from reliable long-horizon grounding agents. Strong performance does not follow from model scale or longer reasoning alone; enabling thinking mode does not necessarily increase AMIGO performance unless it improves valid question selection, evidence use, and stopping decisions. Reliable performance instead requires the combination of fine-grained visual discrimination, effective questioning and stopping, constraint maintenance, protocol adherence, and recovery from controlled noisy feedback. Across these axes, AMIGO exposes qualitatively different failure modes: models may guess correctly without verification-passing evidence, stop early in low-threshold large-gallery failures, spend many turns on invalid or redundant questions, exhaust the question budget without converging, violate the upload-completion boundary, or fail to recover after a single contradictory feedback signal.

The reliability audits support the benchmark as a measurement tool: oracle agreement remains high across both a cross-backbone check and a fashion-expert audit, and the question-violation detector audit indicates that post-upload \texttt{Skip} rates are conservative estimates of invalid-action frequency. Together with the released benchmark data, these diagnostics provide a concrete substrate for future work on protocol-aware multimodal agents. The same protocol can also be instantiated in the other closed-gallery domains illustrated in Figure~\ref{fig:alternative_tasks}. Natural next steps are to broaden the task domains, evaluate closed-source VLMs, and use the collected trajectories for reinforcement learning or preference optimization targeted at valid questioning, belief maintenance, conservative stopping, and recovery from noisy feedback.

\appendix
\clearpage
\section{Failure Cases}

The aggregate results show that current VLMs fail for different reasons: some stop before collecting verification-passing evidence, some continue asking questions but repeatedly violate the protocol, and some struggle to integrate valid constraints into a unique target. We illustrate these behaviors with representative trajectories. All examples use the same task prompt and protocol; images are uploaded in batches of five, and the hidden target is selected from the resulting gallery.

\subsection{Benchmark Task Prompt}

\begin{tcolorbox}[colback=gray!5, colframe=gray!60, title={\textbf{Prompt}}, fonttitle=\small, fontupper=\small, left=4pt, right=4pt, top=2pt, bottom=2pt]
  \textbf{Task:} I will upload several batches of dress options. Please guess which one is my favorite dress and provide its index in the image gallery (starting index: 1).

  \textbf{Constraints:}
  \begin{enumerate}[leftmargin=*, nosep]
    \item Ask questions to gather attribute information about the target dress.
    \item Ask exactly one Yes/No question per turn about one distinguishable feature. I will respond with Yes, No, Unsure (if not confident), or Skip (if you violate the rules).
    \item If you receive Skip, your latest question was invalid and provides no information.
    \item Do not ask about sleeve length, garment length, piece count, color, pattern/prints, age group, size, shoes, necklace, hat, bag, background, or the human model.
    \item Do not enumerate attribute values across turns for the same attribute type (e.g., square neck, V-neck, crew neck).
    \item Do not inspect images one-by-one or reference indices (e.g., ``is your favorite dress the first image?'').
    \item Do not guess until you have narrowed the target down to one image.
    \item If your confidence is low or you find contradictions, keep asking verification questions.
    \item When ready, output: \texttt{My guess of your favorite dress: \#<number>.}
    \item Only start generating questions after you receive \texttt{End of uploading}.
    \item You may ask at most 20 questions to collect information.
  \end{enumerate}
\end{tcolorbox}

\subsection{Qwen3-VL-235B-A22B-Instruct-FP8 failure cases}

\begin{figure}[htbp]
  \centering
  \includegraphics[width=0.85\textwidth]{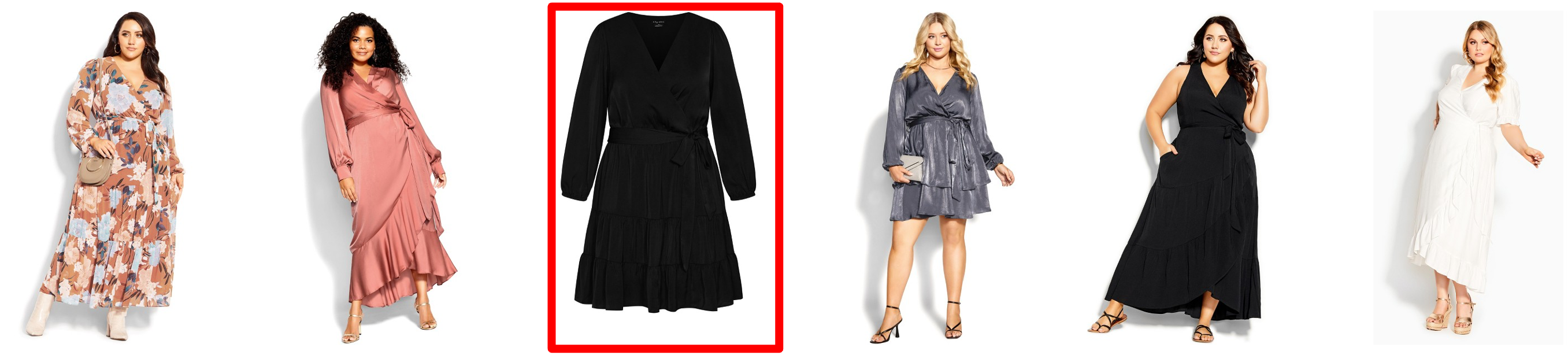}
   \vspace{0.3em}
  \small
  \begin{tabular}{r l >{\raggedright\arraybackslash}p{0.65\textwidth}}
    \toprule
    \textbf{Turn} & \textbf{Speaker} & \textbf{Content} \\
    \midrule
     & \textcolor{green!50!black}{User}   & Prompt + \texttt{<image\_batch\_1>} \\
        & \textcolor{blue}{Model}  & End of uploading \\
          & \textcolor{green!50!black}{User}   & End of uploading + \texttt{<image\_batch\_2>} \\
     1 & \textcolor{blue}{Model}  & Does the dress have a tiered skirt? \\
      & \textcolor{green!50!black}{User}   & Yes \\
    2 & \textcolor{blue}{Model}  & Is the dress made of a shiny or satin-like fabric? \\
      & \textcolor{green!50!black}{User}   & Yes \\
    \rowcolor{yellow!30} 3 & \textcolor{blue}{Model}  & Does the dress have long sleeves? \\
    \rowcolor{yellow!30}   & \textcolor{green!50!black}{User}   & \textbf{\textcolor{orange!80!black}{Skip}} \hfill\textit{\textcolor{red}{Protocol 4: sleeve length}} \\
    \rowcolor{yellow!30} 4 & \textcolor{blue}{Model}  & Is the dress sleeveless? \\
    \rowcolor{yellow!30}   & \textcolor{green!50!black}{User}   & \textbf{\textcolor{orange!80!black}{Skip}} \hfill\textit{\textcolor{red}{Protocol 4: sleeve length}} \\
    5 & \textcolor{blue}{Model}  & Does the dress have a wrap-style front? \\
      & \textcolor{green!50!black}{User}   & Yes \\
    \rowcolor{yellow!30} 6 & \textcolor{blue}{Model}  & Is the dress floor-length? \\
    \rowcolor{yellow!30}   & \textcolor{green!50!black}{User}   & \textbf{\textcolor{orange!80!black}{Skip}} \hfill\textit{\textcolor{red}{Protocol 4: garment length}} \\
    7 & \textcolor{blue}{Model}  & Does the dress have a high-low hemline? \\
      & \textcolor{green!50!black}{User}   & No \\
    \rowcolor{yellow!30} 8 & \textcolor{blue}{Model}  & Is the dress in a solid color? \\
    \rowcolor{yellow!30}   & \textcolor{green!50!black}{User}   & \textbf{\textcolor{orange!80!black}{Skip}} \hfill\textit{\textcolor{red}{Protocol 4: color}} \\
    9 & \textcolor{blue}{Model}  & Does the dress have a ruffled hem? \\
      & \textcolor{green!50!black}{User}   & Yes \\
    \midrule
      & \textcolor{blue}{Model}  & \textbf{My guess: dress \#2} \quad (\textcolor{red}{\ding{55}} Incorrect) \\
    \bottomrule
  \end{tabular}
   \caption{Qwen3-VL-235B-A22B-Instruct-FP8 failure at $\tau=0.8$: prohibited-attribute questions consume four of nine turns before an incorrect final guess.}
  \label{fig:failure_case_1}
\end{figure}

In Figure~\ref{fig:failure_case_1}, the first \textbf{Qwen3-VL-235B-A22B-Instruct-FP8} example occurs in a small but visually demanding $\tau=0.8$ gallery with 6 candidate images. The trajectory mixes useful construction-focused questions about tiering, fabric sheen, wrap-front structure, hem shape, and ruffled details with invalid questions about prohibited attributes. Four of the 9 rounds are returned as \texttt{Skip} for \textbf{Protocol 4} violations involving sleeve length, garment length, and color. Because these turns provide no oracle information, the model spends nearly half of the dialogue budget unproductively and then selects the wrong dress.

\begin{figure}[htbp]
  \centering
  \includegraphics[width=0.85\textwidth]{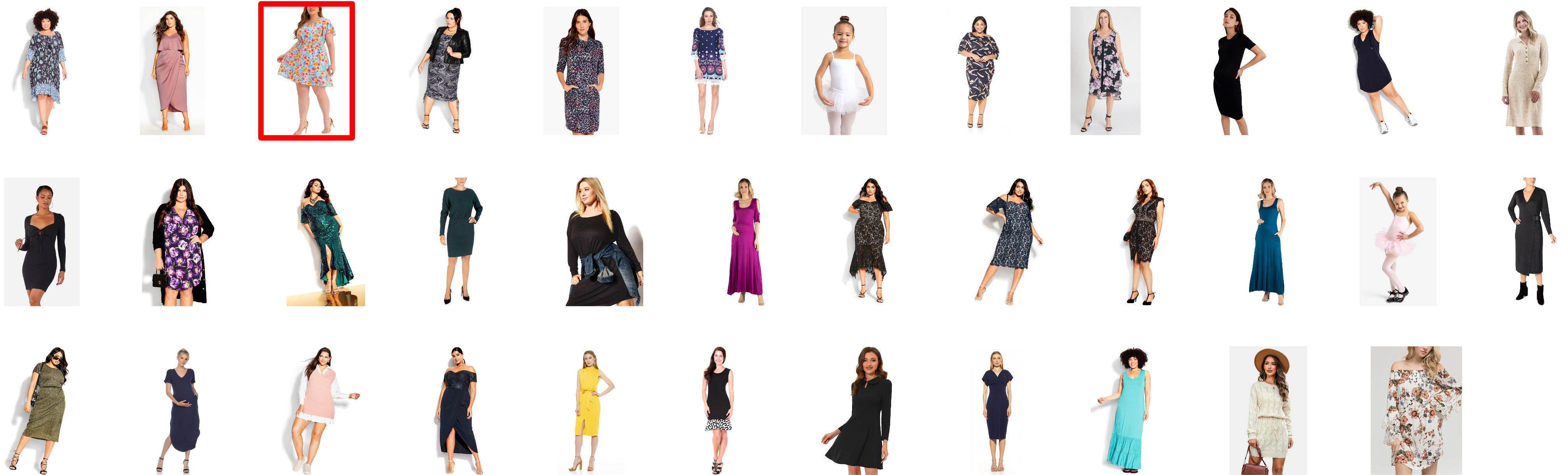}
   \vspace{0.3em}
  \small
  \begin{tabular}{r l >{\raggedright\arraybackslash}p{0.65\textwidth}}
    \toprule
    \textbf{Turn} & \textbf{Speaker} & \textbf{Content} \\
    \midrule
      & \textcolor{green!50!black}{User}   & Prompt + \texttt{<image\_batch\_1>} \\
      & \textcolor{blue}{Model}  & End of uploading \\
      & \textcolor{green!50!black}{User}   & Here is the next batch of dress options. \texttt{<image\_batch\_2>} \\
      & \textcolor{blue}{Model}  & End of uploading \\
      & \textcolor{green!50!black}{User}   & Here is the next batch of dress options. \texttt{<image\_batch\_3>} \\
    \rowcolor{yellow!30}   & \textcolor{blue}{Model}  & Does the dress have a ruffled hem? \hfill\textit{\textcolor{red}{Protocol 10: upload phase}} \\
      & \textcolor{green!50!black}{User}   & Here is the next batch of dress options. \texttt{<image\_batch\_4>} \\
    \rowcolor{yellow!30}   & \textcolor{blue}{Model}  & Does the dress have a lace overlay? \hfill\textit{\textcolor{red}{Protocol 10: upload phase}} \\
      & \textcolor{green!50!black}{User}   & Here is the next batch of dress options. \texttt{<image\_batch\_5>} \\
    \rowcolor{yellow!30}   & \textcolor{blue}{Model}  & Does the dress have a high neckline? \hfill\textit{\textcolor{red}{Protocol 10: upload phase}} \\
      & \textcolor{green!50!black}{User}   & Here is the next batch of dress options. \texttt{<image\_batch\_6>} \\
    \rowcolor{yellow!30}   & \textcolor{blue}{Model}  & Does the dress have a V-neckline? \hfill\textit{\textcolor{red}{Protocol 10: upload phase}} \\
      & \textcolor{green!50!black}{User}   & End of uploading \texttt{<image\_batch\_7>} \\
    1 & \textcolor{blue}{Model}  & Does the dress have a front tie detail? \\
      & \textcolor{green!50!black}{User}   & Yes \\
    \midrule
      & \textcolor{blue}{Model}  & \textbf{My guess: dress \#1} \quad (\textcolor{red}{\ding{55}} Incorrect) \\
    \bottomrule
  \end{tabular}
  \caption{Qwen3-VL-235B-A22B-Instruct-FP8 failure at $\tau=0.3$: premature upload-phase questions disrupt the dialogue before an early incorrect guess.}
  \label{fig:failure_case_2}
\end{figure}

Figure~\ref{fig:failure_case_2} presents the second \textbf{Qwen3-VL-235B-A22B-Instruct-FP8} example, from the broad-gallery $\tau=0.3$ setting, where 35 candidate images are uploaded across seven batches. The main issue appears before the formal questioning phase: the model asks attribute questions while images are still being provided, violating \textbf{Protocol 10}. Once the true upload phase ends, the model asks only one valid question before guessing incorrectly. This behavior directly illustrates Qwen3-VL's broad-gallery early-stop pattern, where very short incorrect traces can coexist with longer verified traces.

\subsection{Qwen3.5-397B-A17B-FP8 failure cases}

In the first \textbf{Qwen3.5-397B-A17B-FP8} example (Figure~\ref{fig:failure_case_qwen35_1}), the model uses the full 20-question budget in a $\tau=0.6$ gallery with 7 candidate images but never produces a final guess. Ten turns receive \texttt{Skip}: the model repeatedly enumerates neckline types after already asking about neckline, asks about prohibited sleeve-length and print attributes, re-enumerates skirt type, and later returns to prohibited garment-length and color questions. The valid turns establish several potentially useful constraints, including smocked bodice, side slit, and fitted waist, but the model does not integrate this evidence into a unique identification. This trajectory shows that conservative interaction length alone is insufficient when the model cannot maintain the protocol or synthesize constraints effectively.

The second \textbf{Qwen3.5-397B-A17B-FP8} example (Figure~\ref{fig:failure_case_qwen35_2}), at $\tau=0.5$ with 14 candidate images, exposes a different version of the same problem. The valid questions identify several target attributes, including wrap-style structure, V-neckline, slit, ruched detail, and high-low hemline. Rather than using these constraints to ask new discriminative questions, the model first asks a prohibited print question and then enters a \textbf{Protocol 5} loop of compound questions that recombine already-confirmed attributes. These invalid turns return \texttt{Skip}, and the model exhausts all 20 turns without a final answer. This failure highlights a distinct reasoning problem: the model retains some local facts from the dialogue, but fails to use them as a state representation for candidate-set reduction.

Overall, these cases connect the aggregate metrics to concrete interaction failures. \textbf{Qwen3-VL-235B-A22B-Instruct-FP8} has threshold-dependent trace lengths: it can stop early in broad galleries, but at stricter thresholds it can also spend many turns on \textbf{Protocol 4} prohibited-attribute questions before an incorrect guess. \textbf{Qwen3.5-397B-A17B-FP8} is more persistent, but often spends its longer traces on \textbf{Protocol 4} prohibited-attribute violations and \textbf{Protocol 5} redundant enumeration. Together, the examples show why AMIGO evaluates more than final correctness: successful performance requires sustained protocol compliance, discriminative question selection, constraint tracking, and evidence-grounded stopping.

\clearpage
\begin{figure}[p]
  \centering
  \includegraphics[width=0.72\textwidth]{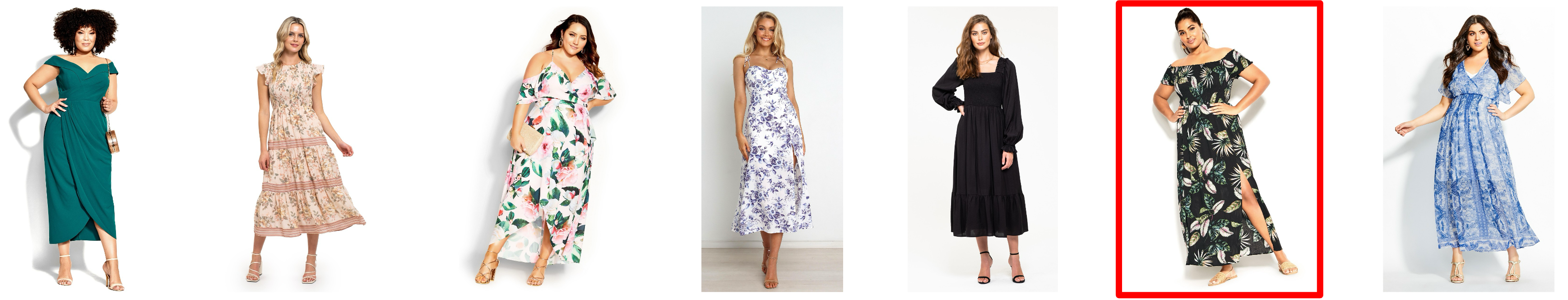}
   \vspace{0.3em}
  \small
  \begin{tabular}{r l >{\raggedright\arraybackslash}p{0.52\textwidth}}
    \toprule
    \textbf{Turn} & \textbf{Speaker} & \textbf{Content} \\
    \midrule
     & \textcolor{green!50!black}{User}   & Prompt + \texttt{<image\_batch\_1>}  \\
           & \textcolor{blue}{Model}  & Understood! I will wait for the 'End of uploading' signal before generating any questions. Once you confirm that all images have been uploaded, I will begin asking Yes/No questions to identify your favorite dress based on its attributes.\\
      & \textcolor{green!50!black}{User}   & End of uploading + \texttt{<image\_batch\_2>}  \\    
    1  & \textcolor{blue}{Model}          & Does your favorite dress feature a V-neckline? \\
       & \textcolor{green!50!black}{User}  & No \\
    \rowcolor{yellow!30} 2  & \textcolor{blue}{Model}          & Does your favorite dress have an off-the-shoulder neckline? \\
    \rowcolor{yellow!30}    & \textcolor{green!50!black}{User}  & \textbf{\textcolor{orange!80!black}{Skip}} \hfill\textit{\textcolor{red}{Protocol 5: enumerating neckline types}} \\
    \rowcolor{yellow!30} 3  & \textcolor{blue}{Model}          & Does your favorite dress have a square neckline? \\
    \rowcolor{yellow!30}    & \textcolor{green!50!black}{User}  & \textbf{\textcolor{orange!80!black}{Skip}} \hfill\textit{\textcolor{red}{Protocol 5: enumerating neckline types}} \\
    \rowcolor{yellow!30} 4  & \textcolor{blue}{Model}          & Does your favorite dress have a sweetheart neckline? \\
    \rowcolor{yellow!30}    & \textcolor{green!50!black}{User}  & \textbf{\textcolor{orange!80!black}{Skip}} \hfill\textit{\textcolor{red}{Protocol 5: enumerating neckline types}} \\
    5  & \textcolor{blue}{Model}          & Does your favorite dress have a wrap-style bodice? \\
       & \textcolor{green!50!black}{User}  & No \\
    6  & \textcolor{blue}{Model}          & Does your favorite dress have a tiered skirt? \\
       & \textcolor{green!50!black}{User}  & No \\
    7  & \textcolor{blue}{Model}          & Does your favorite dress have a smocked bodice? \\
       & \textcolor{green!50!black}{User}  & Yes \\
    8  & \textcolor{blue}{Model}          & Does your favorite dress have flutter sleeves? \\
       & \textcolor{green!50!black}{User}  & No \\
    \rowcolor{yellow!30} 9  & \textcolor{blue}{Model}          & Does your favorite dress have long sleeves? \\
    \rowcolor{yellow!30}    & \textcolor{green!50!black}{User}  & \textbf{\textcolor{orange!80!black}{Skip}} \hfill\textit{\textcolor{red}{Protocol 4: sleeve length}} \\
    10 & \textcolor{blue}{Model}          & Does your favorite dress have a side slit? \\
       & \textcolor{green!50!black}{User}  & Yes \\
    \bottomrule
  \end{tabular}
  \caption{Qwen3.5-397B-A17B-FP8 failure at $\tau=0.6$: repeated P4/P5 violations exhaust the 20-question budget without a final guess.}
  \label{fig:failure_case_qwen35_1}
\end{figure}

\begin{figure}[p]
  \ContinuedFloat
  \centering
  \small
  \begin{tabular}{r l >{\raggedright\arraybackslash}p{0.52\textwidth}}
    \toprule
    \textbf{Turn} & \textbf{Speaker} & \textbf{Content} \\
    \midrule
    \rowcolor{yellow!30} 11 & \textcolor{blue}{Model}          & Does your favorite dress have a floral print? \\
    \rowcolor{yellow!30}    & \textcolor{green!50!black}{User}  & \textbf{\textcolor{orange!80!black}{Skip}} \hfill\textit{\textcolor{red}{Protocol 4: pattern/prints}} \\
    \rowcolor{yellow!30} 12 & \textcolor{blue}{Model}          & Does your favorite dress have a leaf print? \\
    \rowcolor{yellow!30}    & \textcolor{green!50!black}{User}  & \textbf{\textcolor{orange!80!black}{Skip}} \hfill\textit{\textcolor{red}{Protocol 4: pattern/prints}} \\
    13 & \textcolor{blue}{Model}          & Does your favorite dress have a high-low hemline? \\
       & \textcolor{green!50!black}{User}  & No \\
    14 & \textcolor{blue}{Model}          & Does your favorite dress have a straight hemline? \\
       & \textcolor{green!50!black}{User}  & No \\
    15 & \textcolor{blue}{Model}          & Does your favorite dress have a ruffled hemline? \\
       & \textcolor{green!50!black}{User}  & No \\
    \rowcolor{yellow!30} 16 & \textcolor{blue}{Model}          & Does your favorite dress have a flared skirt? \\
    \rowcolor{yellow!30}    & \textcolor{green!50!black}{User}  & \textbf{\textcolor{orange!80!black}{Skip}} \hfill\textit{\textcolor{red}{Protocol 5: enumerating skirt types}} \\
    17 & \textcolor{blue}{Model}          & Does your favorite dress have a fitted waist? \\
       & \textcolor{green!50!black}{User}  & Yes \\
    \rowcolor{yellow!30} 18 & \textcolor{blue}{Model}          & Does your favorite dress have a midi length? \\
    \rowcolor{yellow!30}    & \textcolor{green!50!black}{User}  & \textbf{\textcolor{orange!80!black}{Skip}} \hfill\textit{\textcolor{red}{Protocol 4: garment length}} \\
    \rowcolor{yellow!30} 19 & \textcolor{blue}{Model}          & Does your favorite dress have a maxi length? \\
    \rowcolor{yellow!30}    & \textcolor{green!50!black}{User}  & \textbf{\textcolor{orange!80!black}{Skip}} \hfill\textit{\textcolor{red}{Protocol 4: garment length}} \\
    \rowcolor{yellow!30} 20 & \textcolor{blue}{Model}          & Does your favorite dress have a solid color? \\
    \rowcolor{yellow!30}    & \textcolor{green!50!black}{User}  & \textbf{\textcolor{orange!80!black}{Skip}} \hfill\textit{\textcolor{red}{Protocol 4: color}} \\
    \midrule
       & \textcolor{blue}{Model}          & \textbf{No guess made --- 20-question budget exhausted} \quad (\textcolor{red}{\ding{55}} Incorrect) \\
    \bottomrule
  \end{tabular}
  \caption[]{Qwen3.5-397B-A17B-FP8 failure at $\tau=0.6$ (\emph{continued}).}
\end{figure}

\clearpage
\begin{figure}[p]
  \centering
  \includegraphics[width=0.72\textwidth]{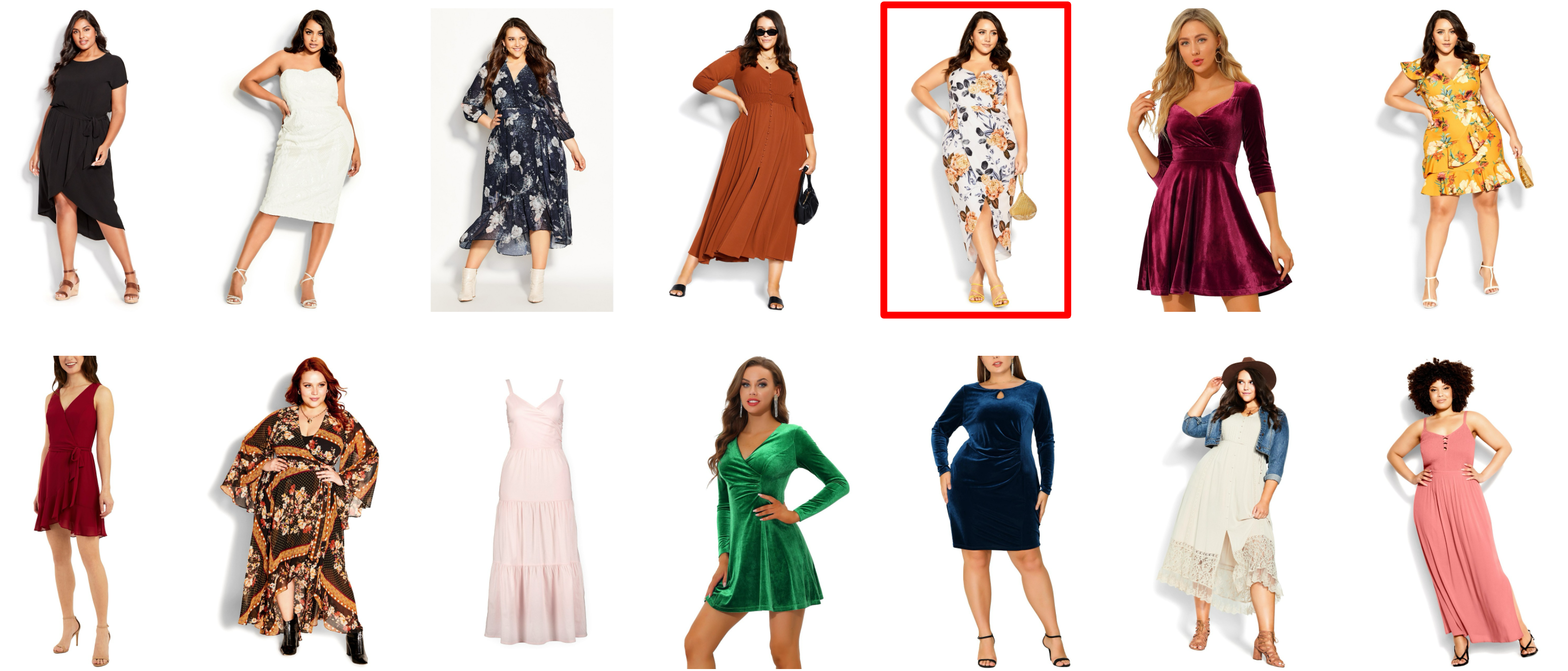}
  \vspace{0.3em}
{\small
\begin{tabular}{@{}r l >{\raggedright\arraybackslash}p{0.52\textwidth}@{}}
  \toprule
  \textbf{Turn} & \textbf{Speaker} & \textbf{Content} \\
  \midrule

   & \textcolor{green!50!black}{User}   & Prompt + \texttt{<image\_batch\_1>} \\
    & \textcolor{blue}{Model}          &  Understood! I will wait for the 'End of uploading' signal before generating any questions. Please let me know when you are ready. \\
     & \textcolor{green!50!black}{User}   & End of uploading + \texttt{<image\_batch\_2>} \\
  1  & \textcolor{blue}{Model}          & Is your favorite dress made of velvet fabric? \\
     & \textcolor{green!50!black}{User}  & No \\
  2  & \textcolor{blue}{Model}          & Is your favorite dress a wrap-style dress? \\
     & \textcolor{green!50!black}{User}  & Yes \\
  \rowcolor{yellow!30} 3  & \textcolor{blue}{Model}          & Does your favorite dress have a floral print? \\
  \rowcolor{yellow!30}    & \textcolor{green!50!black}{User}  & \textbf{\textcolor{orange!80!black}{Skip}} \hfill\textit{\textcolor{red}{Protocol 4: pattern/prints}} \\
  4  & \textcolor{blue}{Model}          & Does your favorite dress have a V-neckline? \\
     & \textcolor{green!50!black}{User}  & Yes \\
  5  & \textcolor{blue}{Model}          & Does your favorite dress have a tiered skirt? \\
     & \textcolor{green!50!black}{User}  & No  \\
  6  & \textcolor{blue}{Model}          & Does your favorite dress have a slit? \\
     & \textcolor{green!50!black}{User}  & Yes \\
  7  & \textcolor{blue}{Model}          & Does your favorite dress have a belt or tie at the waist? \\
     & \textcolor{green!50!black}{User}  & No \\
  8  & \textcolor{blue}{Model}          & Does your favorite dress have a ruched detail? \\
     & \textcolor{green!50!black}{User}  & Yes \\
  \rowcolor{yellow!30} 9  & \textcolor{blue}{Model}          & Does your favorite dress have a strapless neckline? \\
  \rowcolor{yellow!30}    & \textcolor{green!50!black}{User}  & \textbf{\textcolor{orange!80!black}{Skip}} \hfill\textit{\textcolor{red}{Protocol 5: enumerating neckline types}} \\
  \rowcolor{yellow!30} 10 & \textcolor{blue}{Model}          & Does your favorite dress have a sweetheart neckline? \\
  \rowcolor{yellow!30}    & \textcolor{green!50!black}{User}  & \textbf{\textcolor{orange!80!black}{Skip}} \hfill\textit{\textcolor{red}{Protocol 5: enumerating neckline types}} \\
  \bottomrule
\end{tabular}
}
  \caption{Qwen3.5-397B-A17B-FP8 failure at $\tau=0.5$: confirmed attributes trigger a P5 re-enumeration loop and no final guess.}
  \label{fig:failure_case_qwen35_2}
\end{figure}

\begin{figure}[p]
  \ContinuedFloat
  \centering
{\small
\begin{tabular}{@{}r l >{\raggedright\arraybackslash}p{0.52\textwidth}@{}}
  \toprule
  \textbf{Turn} & \textbf{Speaker} & \textbf{Content} \\
  \midrule
  11 & \textcolor{blue}{Model}          & Does your favorite dress have a high-low hemline? \\
     & \textcolor{green!50!black}{User}  & Yes \\
  \rowcolor{yellow!30} 12 & \textcolor{blue}{Model}          & Does your favorite dress have a wrap-style bodice with a V-neckline, a ruched detail, and a high-low hemline? \\
  \rowcolor{yellow!30}    & \textcolor{green!50!black}{User}  & \textbf{\textcolor{orange!80!black}{Skip}} \hfill\textit{\textcolor{red}{Protocol 5: re-enumerating confirmed attributes}} \\
  \rowcolor{yellow!30} 13 & \textcolor{blue}{Model}          & Does your favorite dress have a wrap-style bodice with a V-neckline and a high-low hemline? \\
  \rowcolor{yellow!30}    & \textcolor{green!50!black}{User}  & \textbf{\textcolor{orange!80!black}{Skip}} \hfill\textit{\textcolor{red}{Protocol 5: re-enumerating confirmed attributes}} \\
  \rowcolor{yellow!30} 14 & \textcolor{blue}{Model}          & Does your favorite dress have a V-neckline and a high-low hemline? \\
  \rowcolor{yellow!30}    & \textcolor{green!50!black}{User}  & \textbf{\textcolor{orange!80!black}{Skip}} \hfill\textit{\textcolor{red}{Protocol 5: re-enumerating confirmed attributes}} \\
  \rowcolor{yellow!30} 15 & \textcolor{blue}{Model}          & Does your favorite dress have a V-neckline and a slit? \\
  \rowcolor{yellow!30}    & \textcolor{green!50!black}{User}  & \textbf{\textcolor{orange!80!black}{Skip}} \hfill\textit{\textcolor{red}{Protocol 5: re-enumerating confirmed attributes}} \\
  \rowcolor{yellow!30} 16 & \textcolor{blue}{Model}          & Does your favorite dress have a V-neckline and ruched detailing? \\
  \rowcolor{yellow!30}    & \textcolor{green!50!black}{User}  & \textbf{\textcolor{orange!80!black}{Skip}} \hfill\textit{\textcolor{red}{Protocol 5: re-enumerating confirmed attributes}} \\
  \rowcolor{yellow!30} 17 & \textcolor{blue}{Model}          & Does your favorite dress have a wrap-style bodice and a high-low hemline? \\
  \rowcolor{yellow!30}    & \textcolor{green!50!black}{User}  & \textbf{\textcolor{orange!80!black}{Skip}} \hfill\textit{\textcolor{red}{Protocol 5: re-enumerating confirmed attributes}} \\
  \rowcolor{yellow!30} 18 & \textcolor{blue}{Model}          & Does your favorite dress have a wrap-style bodice and a slit? \\
  \rowcolor{yellow!30}    & \textcolor{green!50!black}{User}  & \textbf{\textcolor{orange!80!black}{Skip}} \hfill\textit{\textcolor{red}{Protocol 5: re-enumerating confirmed attributes}} \\
  \rowcolor{yellow!30} 19 & \textcolor{blue}{Model}          & Does your favorite dress have a V-neckline and a wrap-style bodice? \\
  \rowcolor{yellow!30}    & \textcolor{green!50!black}{User}  & \textbf{\textcolor{orange!80!black}{Skip}} \hfill\textit{\textcolor{red}{Protocol 5: re-enumerating confirmed attributes}} \\
  \rowcolor{yellow!30} 20 & \textcolor{blue}{Model}          & Does your favorite dress have a V-neckline and a ruched detail? \\
  \rowcolor{yellow!30}    & \textcolor{green!50!black}{User}  & \textbf{\textcolor{orange!80!black}{Skip}} \hfill\textit{\textcolor{red}{Protocol 5: re-enumerating confirmed attributes}} \\
  \midrule
     & \textcolor{blue}{Model}          & \textbf{No guess made --- 20-question budget exhausted} \quad (\textcolor{red}{\ding{55}} Incorrect) \\
  \bottomrule
\end{tabular}
}
  \caption[]{Qwen3.5-397B-A17B-FP8 failure at $\tau=0.5$ (\emph{continued}).}
\end{figure}
\FloatBarrier

\clearpage
\section{Oracle and Question-Violation Detector Prompts}

This section reports the oracle feedback prompt, the deterministic P1 format check, and the question-violation detector prompts used during evaluation. Figure~\ref{fig:oracle_feedback_prompt} gives the oracle feedback prompt; Figures~\ref{fig:question_type_verification_prompt} and~\ref{fig:detect_attribute_from_history_prompt} give the two prompt-based violation-detector checks.

\paragraph{P1 malformed-output detection.}
Before applying LLM-based checks, we label empty turns, non-questions, multi-question turns, and non-Yes/No-answerable questions as P1 violations. The two prompt-based detectors then handle prohibited/index content and redundant attributes.

\subsection{Oracle Feedback Prompt}

\begin{tcolorbox}[breakable, colback=gray!5, colframe=gray!60, title={\textbf{Prompt}}, fonttitle=\small, fontupper=\small, left=4pt, right=4pt, top=2pt, bottom=2pt]
\begin{Verbatim}[fontsize=\small,breaklines=true,breakanywhere=true]
You are a fashion expert and understand the fashion items,
attributes and attribute values very well. You will be shown an
image of a dress and asked questions related to the dress in the
image. Please answer the question with "Yes" or "No", or "Unsure"
if you couldn't see the details clearly.
\end{Verbatim}
\end{tcolorbox}
{\captionsetup{type=figure}
\captionof{figure}{Oracle feedback prompt used to answer valid target-image questions.}
\label{fig:oracle_feedback_prompt}}

\subsection{Question Type Verification Prompt}

\begin{tcolorbox}[breakable, colback=gray!5, colframe=gray!60, title={\textbf{Prompt}}, fonttitle=\small, fontupper=\small, left=4pt, right=4pt, top=2pt, bottom=2pt]
\begin{Verbatim}[fontsize=\small,breaklines=true,breakanywhere=true]
You are a Question Type Violation Detector for fashion image Q&A.

Task:
Given one question, output ONLY one token: Yes or No.
- Yes = the question IS a type violation.
- No  = the question is NOT a type violation.

Violation categories (ANY mention => violation => output Yes):
A) Image gallery / image index
- which image, image number, first/second batch, etc.

B) Dress / product attributes (restricted list)
- sleeve length (sleeveless / short sleeve / long sleeve / cap sleeve)
- garment length (mini / midi / maxi / knee-length)
- pattern / print (including themed prints, character prints)
- age group (kids/teen/adult)
- size (S/M/L, numeric sizes)
- color (any color words, solid color, two-tone, monochrome,
  light/dark, bright, pastel, colorful)

C) Presence or type of these objects
- shoes / footwear
- outerwear / outlayer (jacket, coat, cardigan, blazer)
- necklace
- hat
- bag
- background / scene (location/setting/backdrop)

D) Piece count / number of pieces
- one-piece / single-piece
- two-piece / 2-piece / two piece
- three-piece / 3-piece / three piece
- set / matching set / co-ord / coord
- separates / separate top and skirt / two-piece set

Rule:
- If the question asks whether the outfit is one-piece vs
  multi-piece, or mentions set/separates/piece count, it is
  a violation -> output Yes.

Non-violation attributes (examples; NOT exhaustive):
- hemline style (high-low, ruffled hem, tiered skirt, lace
  trim at hem, etc.)
- decoration (ruffles, bows, belts/ties, embroidery/applique,
  trim, embellishments)
- construction details (neckline/collar types, straps, sleeve
  style like puff/balloon/kimono/flutter, silhouette/bodycon,
  panels, bodice details, cuffs)
- material / fabric (lace, sequins, sheen, metallic, sheer, etc.)
- any other attribute not listed in Violation categories

Critical combination rule:
- If a question mentions ANY Violation category AND also mentions
  ANY Non-violation attribute, output Yes.
- If a question mentions ANY Violation category (even alone),
  output Yes.
- Output No ONLY if the question contains ZERO Violation
  categories.


Few-shot examples

Violation -> Yes
Q: Does the dress have white collar?
A: Yes
Q: Does your favorite dress come with a matching hair accessory?
A: Yes
Q: Is the dress designed with a playful or thematic print?
A: Yes
Q: Does the dress have a solid color or a simple pattern?
A: Yes
Q: Is the dress with a floral pattern?
A: Yes
Q: Is the dress designed for kids or teenagers?
A: Yes
Q: Is the dress paired with sandals?
A: Yes
Q: Is the dress worn with a jacket?
A: Yes
Q: Is the dress from the first batch of the image gallery?
A: Yes
Q: Is the dress sleeveless?
A: Yes
Q: Is the background outdoors or indoors?
A: Yes
Q: Is the bag shown a tote or a crossbody?
A: Yes
Q: Does your favorite dress have long sleeves with a slightly
   puffed or voluminous design?
A: Yes
Q: Is your favorite dress a two-piece set with a separate top
   and skirt?
A: Yes

Non-Violation -> No
Q: Does the dress have a tiered skirt?
A: No
Q: Is the dress with high-low hemline?
A: No
Q: Does the dress have a ruffled hem?
A: No
Q: Does the dress have a tie or belt detail at the waist?
A: No
Q: Is the dress made of a shiny or sequined material?
A: No
Q: Is the dress made of lace fabric?
A: No
Q: Does the dress have puff sleeves?
A: No
Q: Does the dress have thin straps?
A: No
Q: Does the dress have an off-the-shoulder design?
A: No
Q: Does the dress have a V-neckline?
A: No
Q: Is the dress form-fitting or bodycon in silhouette?
A: No
Q: Does the dress feature sheer fabric on the sleeves or
   body panels?
A: No
Q: Does the dress have a Peter Pan collar?
A: No
Q: Does the dress have an overlay layer on top?
A: No

Now classify this question (output only Yes/No):
\end{Verbatim}
\end{tcolorbox}
{\captionsetup{type=figure}
\captionof{figure}{Question type verification prompt used to detect protocol violations in fashion image Q\&A.}
\label{fig:question_type_verification_prompt}}

\subsection{Duplicate Attribute Detection Prompt}

\begin{tcolorbox}[breakable, colback=gray!5, colframe=gray!60, title={\textbf{Prompt}}, fonttitle=\small, fontupper=\small, left=4pt, right=4pt, top=2pt, bottom=2pt]
\begin{Verbatim}[fontsize=\small,breaklines=true,breakanywhere=true]
You are given:
- New question (may ask about one OR multiple dress attributes)
- History questions (a numbered list of previous attribute
  questions)

Task:
1) Determine whether ANY fine-grained attribute type mentioned
   in the New question has already appeared in the History
   questions.
2) Output format:
   - If there is a match: output exactly "Yes | <ATTRIBUTE_TYPE>"
     where <ATTRIBUTE_TYPE> is ONE matched fine-grained
     attribute type.
   - If there is no match: output exactly "No"
No other text.

Definition: fine-grained attribute type
A fine-grained attribute type is the most specific attribute
label you can infer from a question.
It is determined by:
  (A) the base item being described (e.g., neckline, hem,
      skirt, bow tie, zipper, waist)
  AND
  (B) the specific aspect/property being asked about (e.g.,
      shape/type, embellishment, location, presence,
      decoration/finish, silhouette, construction, placement)
  AND sometimes
  (C) a required subtype when the aspect is itself multi-type
      (e.g., tiered vs pleated).

Matching rule (strict):
- Two questions match ONLY if they ask about the SAME
  fine-grained attribute type.
- Do NOT match based on shared keywords alone.
- If you cannot confidently assign or match a fine-grained
  attribute type, treat it as NOT a match.

Value rule:
- Different VALUES match ONLY when they are values of the SAME
  fine-grained attribute type.
- Some areas require value-level granularity; treat each named
  style/construction as its OWN fine-grained type.

Process:
1) Extract all fine-grained attribute types mentioned in the
   New question.
2) Extract all fine-grained attribute types mentioned in the
   History questions.
3) If ANY overlap exists -> match.
4) If multiple overlaps exist, choose ONE matched attribute
   type (any one is fine).

Fine-grained distinctions (non-exhaustive):
- NECKLINE_TYPE (V-neck, square neck, round neck, sweetheart,
  etc.)
  != NECKLINE_EMBELLISHMENT (ruffled neckline, lace-trim
  neckline, beaded neckline, etc.)
- Bow tie LOCATION ("bow tie at the waist/shoulder")
  != Bow tie PRESENCE/DESIGN ("has bow tie design")
- Zipper LOCATION ("zipper at the back")
  != CLOSURE TYPE ("zipper closure" vs "button closure")
- Hem SHAPE/HEIGHT ("high-low hemline", "asymmetrical hemline")
  != Hem FINISH/DECORATION ("ruffled hem")
- Style OCCASION/USE ("casual", "formal", "party")
  != named CUT/STYLES (wrap-style, fit-and-flare, shift, tunic)
- Named dress CUT/STYLES are each their OWN fine-grained type
  (no merging):
  * WRAP_STYLE != FIT_AND_FLARE_STYLE != SHIFT_STYLE
    != TUNIC_STYLE

IMPORTANT: Fine-grained skirt rule (do-not-merge)
- "Skirt silhouette" and "skirt construction" are DIFFERENT
  fine-grained attribute types.
- Do NOT match a skirt silhouette question with a skirt
  construction question.
Skirt silhouette (one fine-grained type: SKIRT_SILHOUETTE):
- Values include: flared skirt, pencil skirt, A-line skirt,
  mermaid skirt, etc.
- Different silhouette values still match each other.
Skirt construction (fine-grained types, NOT merged):
- Each construction term is its OWN fine-grained attribute type:
  * TIERED_SKIRT
  * PLEATED_SKIRT
  * GATHERED_SKIRT
  (and similar)
- Different construction terms do NOT match each other.

Waist fine-grained distinction (do-not-merge)
- WAISTLINE_PLACEMENT/SEAM_TYPE != WAIST_DEFINITION/SHAPING
  * WAISTLINE_PLACEMENT/SEAM_TYPE asks WHERE the waistline/seam
    sits (empire waist, drop waist, etc.).
  * WAIST_DEFINITION/SHAPING asks whether the waist is
    emphasized (cinched/defined/nipped-in/belted look).
- These do NOT match each other.

General fine-grained rules (do-not-merge):
1) PRESENCE vs DETAIL
- For any feature X, questions about whether X exists are a
  different fine-grained attribute type from questions about
  X's subtype/style/shape/material/location.
- Therefore: X_PRESENCE ("Does it have X?") does NOT match
  X_DETAIL ("Is X pointed/ruffled/V-shaped/at the waist/etc.?").
Examples (Output = No):
*New: "Does the dress have a collar?"*
*History: 1) "Is the collar pointed style?"*
*Output: No*

*New: "Does your favorite dress feature an asymmetric hemline?"*
*History: 1) "Is there ruffled detailing on the hemline?"*
*Output: No*

2) STRUCTURE/SHAPE vs EMBELLISHMENT/DECORATION
- For any garment area A (neckline/hem/skirt/etc.), questions
  about A's structural shape/type are a different fine-grained
  attribute type from questions about A's embellishment/trim/
  decoration.
- Therefore: A_TYPE/SHAPE (e.g., V-neck, square neck, high-low
  hemline) does NOT match A_EMBELLISHMENT (e.g., ruffled
  neckline, lace-trim neckline, ruffled hem).
Examples (Output = No):
*New: "Does the dress have a V-neckline?"*
*History: 1) "Does the dress have a ruffled neckline?"*
*Output: No*

3) Exact repeat vs new-detail exception:
- Output "Yes" when the New question repeats an already-
  discussed fine-grained attribute type/value and adds no new
  fine-grained detail.
- Output "Yes" when the New question asks the same fine-grained
  attribute type again with a different value.
- Output "No" only when the New question repeats an exact or
  nearly exact old attribute value as context AND adds a new
  fine-grained modifier/detail including decoration, trim, seam
  detail, placement detail, material, count, measurement, or
  other specification details.
- If the added modifier is only a refinement of the same
  fine-grained attribute type, output "Yes".
Examples:
*New: "Does the dress have a tiered skirt?"*
*History: 1) "Does the dress have a tiered skirt?"*
*Output: Yes*

*New: "Does the dress have a tiered skirt with ruffle details
       on the seams?"*
*History: 1) "Does the dress have a tiered skirt?"*
*Output: No*

*New: "Does the dress have a V-neckline with lace trim?"*
*History: 1) "Does the dress have a V-neckline?"*
*Output: No*

*New: "Does the tiered skirt have 2 tiers?"*
*History: 1) "Does the dress have a tiered skirt?"*
*Output: No*

HIGH PRIORITY skirt rule (do-not-merge):
- Skirt SILHOUETTE/SHAPE and skirt CONSTRUCTION/LAYERING are
  different fine-grained attribute types.
- "flared skirt" is SKIRT_SILHOUETTE (shape).
- "tiered skirt" is TIERED_SKIRT (construction/layering) and
  must NOT be treated as a silhouette.
- Therefore: "flared skirt" does NOT match "tiered skirt".
Examples:
*New: "Does the dress have a flared skirt?"*
*History: 1) "Does the dress have a tiered skirt?"*
*Output: No*

*New: "Does the dress have a flared skirt?"*
*History: 1) "Does the dress have a pencil skirt?"*
*Output: Yes*
\end{Verbatim}
\end{tcolorbox}
{\captionsetup{type=figure}
\captionof{figure}{Detect attribute from history prompt used to identify duplicate fine-grained attribute questions.}
\label{fig:detect_attribute_from_history_prompt}}
\clearpage

\bibliographystyle{unsrt}
\bibliography{ref}

\end{document}